**Highlights**

- Towards a robust and accurate real-time robotics grasping

- A custom pipeline to coordinate submodules to form a 3D object recognition system

- Performance achieves high metrics score 97.37% (5cm5deg) and 99.37% (ADD)

- A large relative improvement in detection metrics 5cm5deg (62%) and ADD (52.48%)

- A large average running time improvement of 47.6%

# 6D Pose Estimation with Combined Deep Learning and 3D Vision Techniques for a Fast and Accurate Object Grasping

Tuan-Tang Le [1], Trung-Son Le [1], Yu-Ru Chen [1], Joel Vidal [1,2], Chyi-Yeu Lin [1,3,4,*]

[1] Department of Mechanical Engineering, National Taiwan University of Science and Technology, Taipei 106, Taiwan;

[2] Computer Vision and Robotics Institute, University of Girona, 17003 Girona, Spain

[3] Taiwan Building Technology Center, National Taiwan University of Science and Technology, Taipei 106, Taiwan

[4] Center for Cyber-Physical System, National Taiwan University of Science and Technology, Taipei 106, Taiwan

* Correspondence: jerrylin@mail.ntust.edu.tw

**Abstract:** Real-time robotic grasping, supporting a subsequent precise object-in-hand operation task, is a priority target towards highly advanced autonomous systems. However, such an algorithm which can perform sufficiently-accurate grasping with time efficiency is yet to be found. This paper proposes a novel method with a 2-stage approach that combines a fast 2D object recognition using a deep neural network and a subsequent accurate and fast 6D pose estimation based on Point Pair Feature framework to form a real-time 3D object recognition and grasping solution capable of multi-object class scenes. The proposed solution has a potential to perform robustly on real-time applications, requiring both efficiency and accuracy. In order to validate our method, we conducted extensive and thorough experiments involving laborious preparation of our own dataset. The experiment results show that the proposed method scores 97.37% accuracy in 5cm5deg metric and 99.37% in Average Distance metric. Experiment results have shown an overall 62% relative improvement (5cm5deg metric) and 52.48% (Average Distance metric) by using the proposed method. Moreover, the pose estimation execution also showed an average improvement of 47.6% in running time. Finally, to illustrate the overall efficiency of the system in real-time operations, a pick-and-place robotic experiment is conducted and has shown a convincing success rate with 90% of accuracy. This experiment video is available at https://sites.google.com/view/dl-ppf6dpose/.

## 1. Introduction

3D object recognition has ever attracted the attention of computer vision research community. A robust algorithm integrated to a robotic system can liberate a versatility of labor tasks. However, such an algorithm possessing both sufficient accuracy and performance is yet to be found. In addition, most vision systems depend greatly on perception devices which, regardless of recent technology advancement, still undergoes new development trends. In line with the trend of vision-based robotics system development, this research is developed towards industrial and service robotics applications. In particular, the major interest of this research is real-time pick-and-place performed by a service robot with a robot arm which demands a solution with higher precision and shorter execution time, inclusive of scenarios where it is integrated in a robotic system for a successful pick possibly followed by subsequent precision-based tasks. If the posture of the object in the robotic hand is not precise enough for conducting a direct subsequent operation, such as an insertion, the robot will need to conduct an additional operation aiming to measure the precise posture of the object in hand before such precision-based operation. For both industrial and service robots, if they are expected to perform object recognition and grasping and then subsequent accuracy-based operation, the real-time object grasping with sufficient accuracy is an essential task. There are cutting-edge 6D pose estimation algorithms, but the accuracy and time efficiency cannot meet the need of a real-time accurate grasping which can support subsequent precise operation.

By the early 2000s, much progress had been made to solve the long-term challenges of 6D object pose estimation in computer vision. Before depth sensors became popular, research generally relied on 2D images. Aside from the more popular keypoint-based methods [1], others relied on edge information or its more primitive form, image gradients. Such research explored a different approach to better account for lighting change or complex backgrounds. In order to accomplish 3D recognition or 6D pose estimation from 2D images, multi-view methods were proposed by a large number of researchers, some of whom constructed a hierarchical structure of the views. One early work by [2] initiated a measure to quantify the similarity between views of object silhouettes and a hierarchical structure built from groups of similar views. Markus Ulrich et al. [3] advanced hierarchical graphs to practical machine vision applications where the similarity measure was extensively studied and claimed to have robustness in lighting change and cluttered backgrounds [4,5]. A space autonomous application also found benefits in edge-based object recognition [6]. The author trained the hierarchical view-graph by using an unsupervised clustering technique based on Affinity Propagation and the aforementioned similarity measure. To match an input image with the trained views, he utilized particle filtering, which has a better computing cost and a more straight-forward formulation than searching or optimization techniques.

Several multi-view strategies heuristically search the templates for a match or reduce computation in a coarse-to-fine pyramid manner [3,7,8]. In many cases, views are mostly collected sparsely and partially surrounding an object-centered coordinate system where the object is placed at the origin and the camera locates at certain angle intervals [3,9,10]. Others parameterized the camera placement by points on a view sphere, which can be defined at different granularity levels by recursively subdividing an icosahedral or using angle intervals with image pyramids [7,8].

Hinterstoisser's research [11] combines multiple edge-based templates, normalized to achieve viewpoint invariance, with a classifier to improve processing time. His other research [12,13] explores gradient-based similarity measures as an extension to histogram-of-gradient (HOG) method. The measures successfully exploit edge pixel orientation as an additional feature to reduce false positive detection. The approach initially processed solely RGB images but later added surface normal as another depth sensor modality. Some recent research [14,15] has been inspired by these multiview methods.

With the release of multiple low-cost depth sensors including Microsoft Kinect in 2010, followed by Asus Xtion in 2011 [16], an abundance of research focused on feature-based methods and point cloud data. In general, there were mainly two approaches. The first approach describes local surface characteristics of the object using local features. The common workflow of this approach, as described in Yulan Guo et al.'s survey [17], comprises a 3D keypoint detection, local surface feature description, and surface matching. The survey focused on 8 keypoints detection methods and comprehensively discussed a total of 29 keypoint detection and 38 feature descriptors. There are other 4 keypoints descriptors which were reviewed by Bronstein et al. [18] along with other 7 feature descriptors. The second approach describes the whole 3D object using a set of global features such as viewpoint feature histogram, geometric 3D moments, shape distribution [19–21].

While global features seem to show better performance on recognizing objects with few or repeated local surface characteristics, local features show better performance against background clutter and occlusion. Aside from the two main approaches, methods combining characteristics of both global and local approaches [22,23] have been presented. Notably, several works on Point-pair features (PPF) have yielded significant results [22–27]. However, PPF as well as other feature-based and templated-based approaches, require an input searched target and the result is based on a threshold score. This scheme is not efficient for simultaneously detecting multiple object classes since the algorithm has to be evoked iteratively and, therefore, resulting in a linear complexity. Besides, PPF processing time increases exponentially with respect to the number of model and scene points, although sampling solutions are applied to decrease this effect, local surface details are also sacrificed.

In the early 2010s, the community witnessed a transition from solutions relying on the popular machine learning in a non-exhaustive list including Conditional Random Fields [28], Deformable Part Modeling [14,29,30], Random Forest [31] to the ones relying on Deep Neural Networks [15,32–36]. In 2012, Krizhevsky et al. [37] made a breakthrough by successfully implementing a deep Convolutional neural network within a GPU and won the ImageNet challenge far ahead of the 2nd-place winner. By then, the computer vision community was progressing object recognition solutions at a pace never seen before by taking advantage of this new tool, deep neural networks. A well-known mainstream research was RCNN, Fast-RCNN, Faster-RCNN, Mask R-CNN [38–41] which brought 2D object recognition quickly to versatile practical applications.

Among the first researchers to use Convolutional Neural Network (CNN) for pose estimation was Gkioxari et al. [32]. A derivation of a popular 2D recognition R-CNN was deployed to estimate human pose using keypoints and detect action accordingly. Alternatively, Crivellaro et al. [33] estimated the pose of multiple parts of a target object with two consecutive CNNs. The first network generates several image patch hypotheses for each part, which is fed to the second network to obtain the reprojections of the 3D control points of the part. Each part's pose is resolved by 2D-3D correspondence. This enables the scheme to perform robustly in instances of high occlusion, even in cases of just one visible part. The descriptor-based method was renovated by Wohlhart et al. [15], where the descriptor is trained by a CNN to learn both objects' identity and 3D pose. The author illustrates an example of a low-dimensional descriptor trained on multiple views and objects in a 3D manifold plot. Interestingly, the network could learn a descriptor that is well separated between object classes and views of the same class.

Despite the rising number of deep learning research topics, 3D object recognition has not received sufficient attention. Recent algorithms applying deep learning have not benchmarked against strict performance metrics such as BOP [42]. Practical applications demand stricter working conditions such as bin picking, lighting change, scenes including multiple object classes, or a faster per-frame processing speed. Keypoint and pose regression e.g. [15,34] degrade quickly when there is strong occlusion. Fundamentally, the deep learning framework has not solidly resolved theoretical proof of its performance, which leaves failures without accountability.

Some recent remarkable researches on deep learning-based 3D object recognition include PoseCNN [43] and PVNet [44]. PoseCNN uses a two-stage network to first extract a series of feature maps, then followed by an embedding stage to perform 3 simultaneous tasks: semantic labelling, object 3D translation estimation, and object 3D rotation regression. Along with the pixel-wise labeling, the semantic segmentation in the first network branch also provides object bounding-boxes by using a post-processing on the segmentation results. Another network branch estimates pixel-wise keypoint voting maps which are subsequently processed in a Hough-voting manner to derive the object center. The two network heads' outputs are combined to complete an ROI. The ROIs are piped to the pose regression head to predict object 3D rotation. In contrast to PoseCNN, PVNet architecture has a design inclined towards the end-to-end training fashion with the exception of the keypoint voting post processing. As can be seen in Pose-CNN architecture, the mutual piping of network branches complicates the design and implementation. Instead of PoseCNN's complex design, PVNet accomplishes both two tasks including segmentation of all the object classes and voting map regression for all the keypoints with a single fully convolutional network. In another stage, the output voting maps are processed in a RANSAC-like manner to return all keypoint locations. At last, perspective-n-point method is applied to the collected keypoints to deduce the object poses. These state-of-the-art algorithms and most public benchmark frameworks aim at general computer vision challenges that remain far from practical applications.

In this context, the current achievements using deep neural networks are not capable of resolving precise 3D object pose. Moreover, despite being a state-of-the-art algorithm in multiple benchmarks [8,42], the PPF pose estimation has a computing limitation when encountering a multi-class or a dense cloud scene. This computing cost is governed by a global search in scene point-cloud, or in case of [22], an additional heuristic pre-processing to filter the unlikely regions. In both cases, complex scenes e.g. bin-picking… or an unoptimally tuned depth camera can introduce False-Positive votes from the scene cloud points.

The most similar approach to our work was Konig et al. [54] where the author also proposed to use an instance segmentation network in combination with PPF as a main resolution for the processing speed issue. Alongside the hybrid scheme, they also employed a new network training technique with synthetic PBR images and an automatic mechanism to switch the segmentation networks to favor the validation dataset for better performance. Their method claimed to improve the speed 12 times their past implementation [23] and was the best fast method in BOP Challenge 2020. In 2018, Chen et al. [55-56] attached a faster R-CNN [40] 2D pre-recognition stage to PPF to improve the overall 3D recognition performance in a framework that also integrates direct grasp with perspective-n-point, visual servoing. However, the author experienced that the approach does not meet the required performance for accurate robotics applications (interested readers can refer to [55-56] at our project website given in the abstract)

We propose a system integration that can efficiently coordinate a 2D object recognition and a 6D pose estimation algorithms. We show that with proper fine-tuning, this approach can significantly improve the effectiveness and bridge the possibility to real applications. The proposed method attached a state-of-the-art deep learning 2D pre-recognition stage prior to 3D pose estimation to create a 3D object recognition solution. The approach facilitates sole pose estimation with recognition capability and also improves its performance. We built our solution upon the well-known Mask R-CNN framework to first perform object recognition and extract the target bounding box. The extracted bounding-box is used to crop the relevant point cloud therefore helps to reduce the possible False-Positive votes in all scene cloud points, and feed to the PPF pose estimation framework, which inherits the work of Vidal et al. [22] a renowned first-place winner of the BOP competition. The initial object recognition also triggers the PPF algorithm with a supervised signal for the model identity to be found in the scene cloud.

In short, our contribution can be briefly listed as: First, we proposed a custom workflow involving a mobile vision and a one-stage target-approaching move that significantly improves the system performance in practical applications. Second, we integrate, coordinate smoothly both frameworks whose implementations are realized by distinct development foundations by using a client-server interface, and build up a larger 3D object recognition system. Thirdly, a dataset with ground-truth is prepared laboriously and used to benchmark the constituent 2D recognition and 3D pose estimation frameworks. Lastly, to demonstrate the overall system performance in a real scenario, a robotics pick-and-place experiment is conducted and reported with a persuasive success rate.

In the following sections, related work is presented first. The 2D object recognition with instance segmentation Mask R-CNN is described in section 3.1 along with its network architecture. In the following section 3.2, 6D pose estimation PPF is explained along with our workflow to collect object models to prepare for the offline training phase. Section 3.3 represents their integration with a custom workflow and a client-server based interface. Lastly, the solution's performance will be demonstrated by elaborative experiments of both frameworks using their respective evaluation pipelines.

## 2. Related work

In general, the deep-learning based 3D object recognition proposed in the last few years can be categorized into ones that regress directly to 3D pose and ones that regress to keypoints. The latter requires another step to solve 2D-3D correspondence. With the use of deep learning, these algorithms leverage the mostly robust 2D recognition performance with little additional cost for object pose computation. Therefore, they can easily achieve a nearly real-time detection rate which was rare in previous pose estimation methods. Also, deep learning schemes introduce the possibility that 6D pose can be estimated directly from 2D input images, which explains the rising number of publications leaning towards single RGB image pose estimation lately.

Belonging to the first category, in SSD-6D Wadim Kehl et al. [45] extend the 2D object detection single-shot detection concept to 6D pose estimation. The authors trained the neural network by rendered view-sphere images of synthetic 3D models. Given an input RGB image, the network extracts multi-scale feature maps and propagates them through convolution layers to detect bounding-box, object class, a viewpoint identity and an in-plane rotation. Later this output is used to derive the object pose. The method can reach a 10Hz detection rate. In the same category, the work of deep-6D pose represents a common practice where 2D object recognition is deployed as the main pipeline. The detector output was introduced with an additional branch using a fully convolutional network for pose regression. For this case, the pose is interpreted as a 4-element vector with 3 elements representing pose and an estimated Z-depth of the target 3D bounding-box center. To compute the final translation, Z-depth along with the predicted 2D bounding-box coordinates are used with pin-hole camera formulation. Because the object pose relies on the bounding-box center regression, applications requiring accurate pose are not suitable for this method.

However, it is known that for keypoint regression methods, the estimation accuracy degrades quickly when the target is occluded. The same issue occurs in direct pose regression. While the deep learning solutions provide compelling results, a theoretical account of their failures or degradation is lacking. In this work, we employ a hybrid approach that leverages both the contemporary high-performance deep-learning based recognition and a more classic state-of-the-art pose estimation on point cloud input. To coordinate between the frameworks, we design a custom workflow and a client-server interface. This also provides a possibility for future modularization of our system in a larger framework.

In terms of implementation, our system is equipped with two cameras, one is fixed as a global top-view camera, the other is attached to the end-effector of a 6DoF robot arm. In this setup, we assume that whether the eye-in-hand camera is attached to a mobile robot platform or a robot manipulator, the robot location is within a known space that the global camera can presumably cover in advance. Instead of aiming to solve the general and common concept of 3D object recognition problem as in [22,25,42-44,54-56] where the camera perspective to the objects is arbitrary, our work differentiates in a system design and workflow viewpoint where the additional top-view camera and the robot approach move bringing the eye-in-hand camera closer provide leverage to confine both cameras' working distance. In specific, the top-view camera location relative to the object work plane is among the preset parameters. Likewise, the robot approach move can also be automated. In the experiments, we will show that these presets are crucial parameters to bridge the performance gap to accurate robotics applications.

## 3. Method

Figure 1 illustrates the overall proposed architecture of our system. The architecture comprises two main stages. In the first stage, object detection is performed using an instance segmentation network. A list of targets with the corresponding confidence scores, bounding box values and masks will be collected and saved in the database. At each cycle, the manipulator brings the second camera close to the target whose 6D pose is estimated in the second stage. This second stage contains seven steps. For this system, we propose using the pose estimation method presented in [22], including an additional point cloud removal based on the localization result from the deep learning network with target approaching to ensure maximum accuracy from the sensor data.

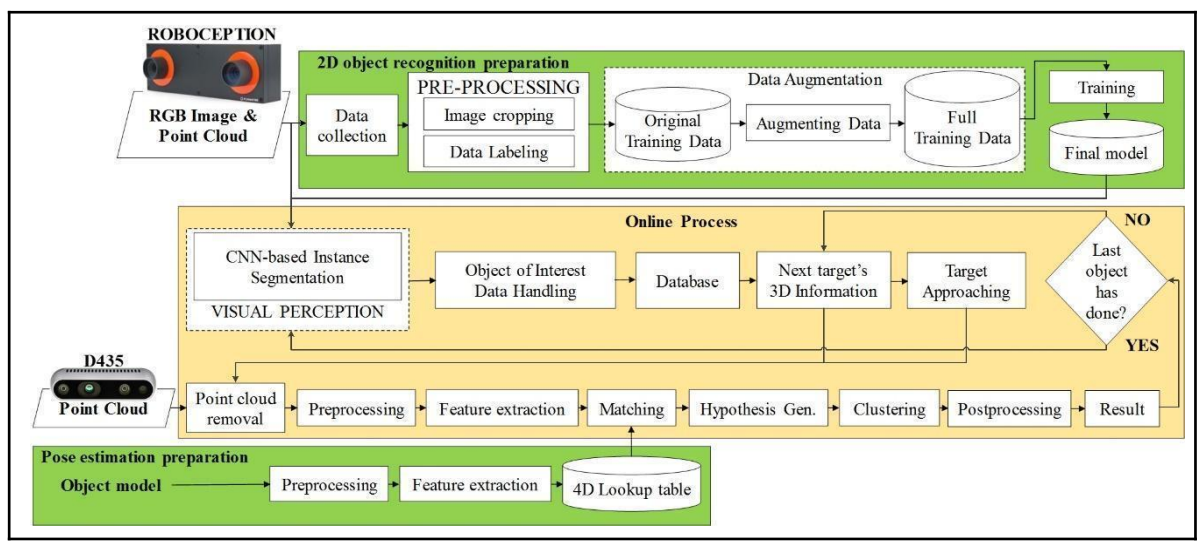

**Figure 1.** Architecture of the proposed system.

*3.1. Object detection*

Localizing the target in the 2D image before mapping to the 3D environment is the first step of the system. In this work, a variant of the Mask R-CNN [41] network provided by Matterport [46], which can detect the objects, label all the shapes in an image, and identify their locations down to the pixel level, is used to initially detect object 2D locations. Other popular networks, e.g. SSD [47], YOLO [48], etc., with a similar 2D detection capability can be considered as alternative candidates.

The network main components include: a backbone, a Feature Pyramid Network (FPN), a Region Proposal Network (RPN), and the three output heads for recognition, bounding box regression, and mask prediction. For clarity and referencing, we briefly overview the network architecture, its components and operational functions. The pipeline starts with preprocessing the input images by adding zero padding to the top and bottom and giving these images a size 960 x 960 fixed squared resolution. An input image is fed to a pre-trained backbone network. In our case, the backbone is a FPN combined with ResNet-101, also referred to as ResNet-FPN. Figure 2 represents the overall network architecture of the Mask R-CNN, which is used in this study.

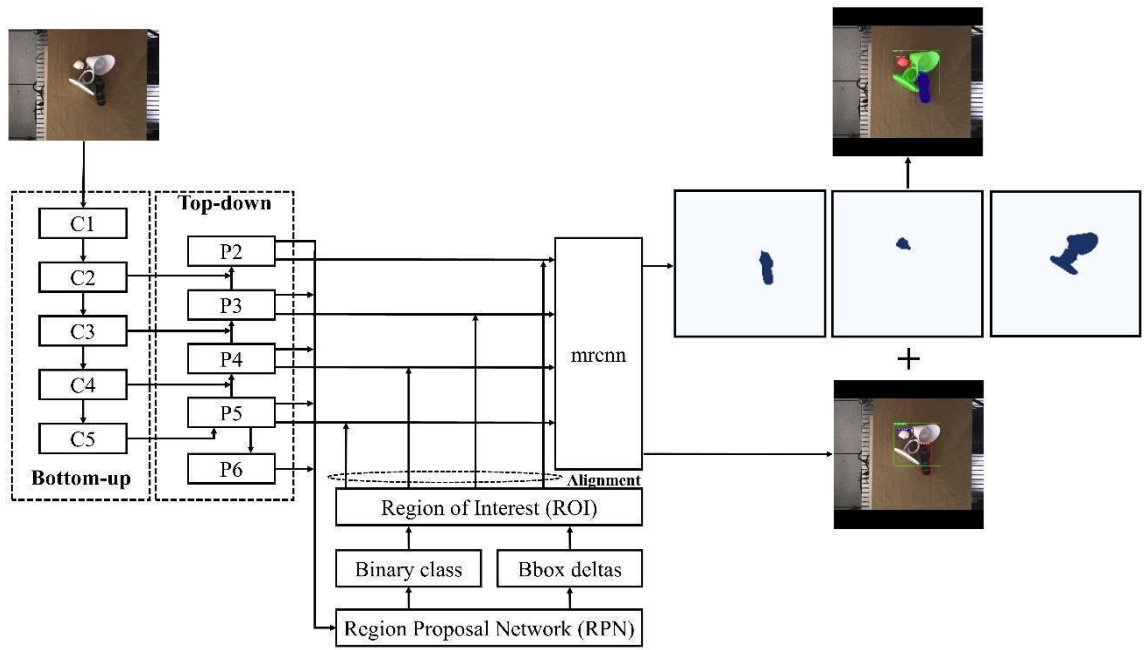

**Figure 2**. Mask R-CNN architecture for object detection.

FPN inherits the classic concept image pyramid, which is a multi-scale representation of the feature maps with the predictions applied independently at each scale. To realize its implementation with a neural network at a low cost, it reuses the available ResNet-101 backbone feature maps {C1, C2, C3, C4, C5}. FPN is constructed from two pathways. In the bottom-up pathway, an FPN receives a list of sequentially scaled feature maps from the input image to the highest-level feature map (C5). In the top-down pathway, these input feature maps are used to build the image pyramid as a series involving upscaling the smaller feature maps and summing them with the similar scale convoluted feature map from the bottom-up pathway. For instance, P4 is generated by upscaling P5 and summing it with C4 which has already been preprocessed by a 1x1 convolution. The commonly used upscaling operators are nearest neighbor or bilinear with a scaling factor of 2.

The backbone network is then placed on top of a RPN which aims to provide the "attention" mechanism and guides the overall network where to look for the targets. The RPN efficiently reuses the feature maps of the detector with a few added convolutional layers. An RPN slides a base 3x3 convolutional operator over a dense grid of 1-2 pixels providing a shared feature map for two branching 1x1 convolutional layers, namely *bbox* and *cls*. The first branch *cls* predicts a binary object/non-object classification returning a 2-tuple, whereas the second branch *bbox* performs the bounding-box regression returning a 4-tuple. With the number of proposals at each point *n*, total output size is *2n* and *4n* for the branch *cls* and *bbox* respectively. When a FPN is integrated to a RPN, the RPN head is applied independently on each scale of the feature pyramid. The proposals are commonly referred to as *anchors*. To differentiate detection bounding-box ratios, the *anchors* are preset at the following ratios {0.5, 1, 2}. Since the image pyramid already has a multi-scale representation, these *anchors* are fixed at a single scale when applied for each pyramid level. The combination of the three ratios and five pyramid scales results in a total of 15 *anchor* shapes.

The use of proposals results in excessive overlapping of bounding-boxes. Therefore, a filtering of these proposals by detection scores is first applied to retain the top 6000 anchors. The remaining anchors are applied with a non-maximum suppression (NMS) having a training proposal count of 2000, an inference count of 1000, and IoU threshold of 0.7.

*3.2. 6D Pose Estimation*

6D pose estimation is the second integral part of the whole pipeline. The first-stage object recognition is fed to a Point Pair Feature voting approach to estimate its 6D pose. PPF was first introduced in 2010 [23] as a mixed global-local feature-based approach. Alongside with template matching, machine learning methods, the feature-based category is known as one of the current state-of-the-art processes for 6D pose estimation using the RGB-D data. The PPF voting approach models the objects' surface globally and performs local matching using sparse descriptors between pairs of points. Each pair of two points on the model surface can form a relationship with each other to create a four-dimensional feature from the relative angles between the points' two normal vectors, a directional vector and a relative distance value. A simple representation of this four-dimensional feature is shown in Figure 3.

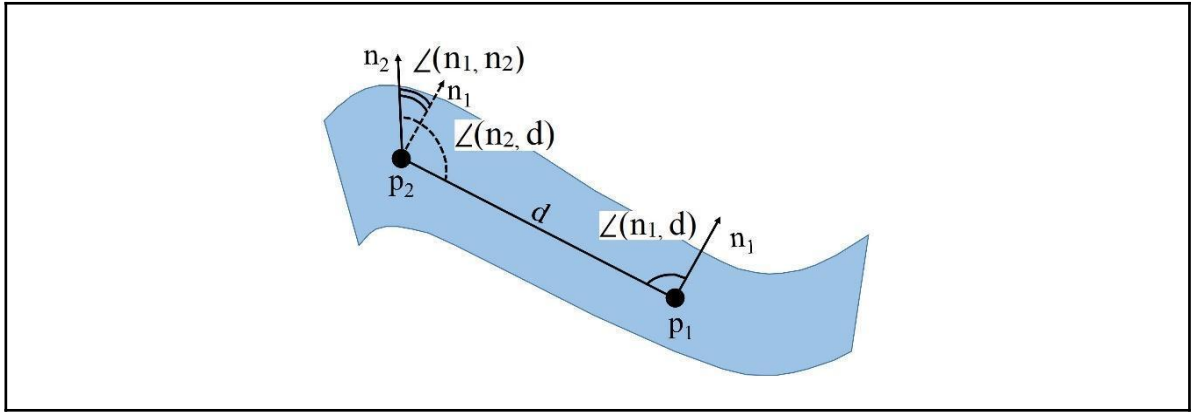

**Figure 3**. The Point Pair Feature demonstration for a model's point pair (p1 , p2).

In Figure 3, we took two points (p1 and p2) as an example to define a point pair features on the surface of an object. Each point will come with their respective normal vector (n1 and n2). The two points can also be used to compute a relative distance (d) and a directional vector pointing from one point to the other. Formally, the four-dimensional feature is defined by equation 1,

$$F(p1, p2, n1, n2) = [||d||, \angle(n1, d), \angle(n2, d), \angle(n1, n2)]^T \quad (1)$$

The distance d is the first element of the four-dimensional feature, followed by three angles of the vector pairs grouped by the two normal vectors and a directional vector, providing three more elements.

As shown in Figure 1, the PPF algorithm is illustrated with two stages: modeling and matching in operational terms or, stage-wise, they can be referred to as the offline and online processes. On the offline process, with the object mesh surface information, the global PPF features are modelled and quantized to fill a 4D lookup table and create the model feature space. In the online process, a matching is executed between the computed point-pairs in a scene and the quantized feature from the model feature space.

The greatest frequency of correspondences supporting a single pose will be considered as the best prediction. In order to reduce the complexity of the voting process, a local coordinate will be used to group the poses within a two-dimensional space efficiently. Therefore, each scene pair will vote for only one position in the two-dimensional table. Finally, all the possible predictions achieved from different scene points are clustered to provide the final hypothesis.

Among many improved methods based on the original PPF algorithm, work from Vidal et al. [22] showed a robust performance on a wide range of 3D objects. Their approach was evaluated against a comprehensive dataset with 68 different types of objects with more than 68,000 scenarios. The stability of this PPF version was once again presented by testing according to the extensive BOP benchmark [42], which outperformed the other 14 state-of-the-art solutions on the same dataset.

A variation of the method presented in [22] will be implemented in this work. Our PPF version is extended with a point cloud removal process. In addition, the combination between the instance segmentation network and the variant PPF version introduces a significant improvement to the system computing complexity by avoiding a function calling iteration when multiple object models in the database are required to match with the target point cloud in multiple object class pose estimations.

*3.3. 6D Pose Estimation Integration for 3D object recognition*

As previously mentioned, the main goal of this work is to enhance the 6D pose estimation of known objects. In this work, we propose a combination of 2D object detection followed by a 2D-3D mapping of this 2D localization to a 3D location to approach the target object and finally run the 6D pose estimation algorithm to derive the 6D pose. Our system is implemented by using a client-server communication model under a custom design pipeline as a protocol to connect two frameworks originally developed under different programming languages.

In this study, a client-server model was used to leverage the popularity of Python in deep learning development and its power as a prototyping and system integration programming language. On the other side, C++ is popular and performance-oriented towards more classic computer vision development. In this work, the Mask R-CNN instance segmentation framework and pose estimation PPF are implemented using Python and C++ languages respectively. The interface between these two functions can be achieved by multiple approaches including code migrating (either from C++ or Python), module wrapping (Cython, Boost Python…), and even using external memory as an intermediate data storage. The Robot Operating System (ROS) architecture inspired the client-server scheme. However, without relying on ROS as a dependency, which might come along with additional ROS protocols and interfacing syntax to the development cost, we chose to build our own interface based on the structure of XMLPRC. This custom interface provides a flexible and simplified protocol to have a timely and responsively functional vision and robotic system as a whole. Again, the client-server protocol is in charge of the connection and coordination between these two frameworks.

A client application could also use Python to evoke the object recognition network and to interact with external devices or machines such as a 6 DoF manipulator, a vacuum suction pad or a gripper in cases where these peripherals are needed. On the server-side, the Roboception and D435 cameras along with the PPF algorithms are written by C++. Figure 4 shows the overall communication pipeline between these two frameworks.

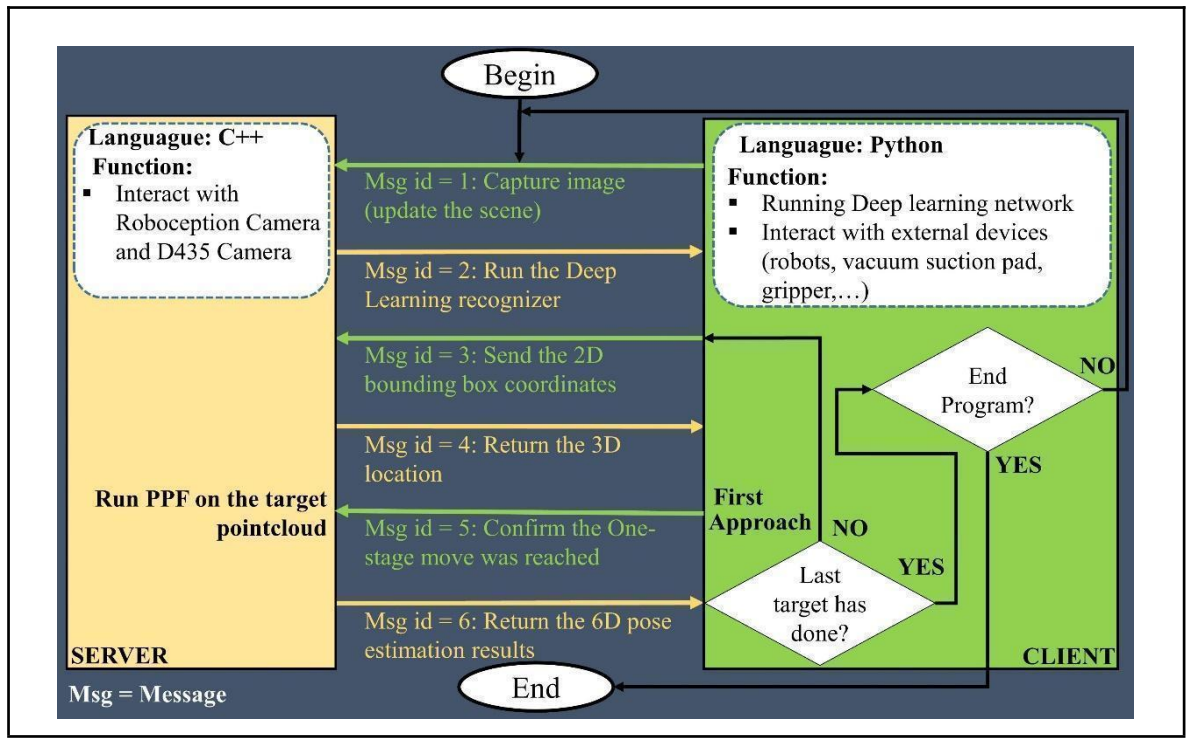

**Figure 4**. The flowchart describes the interaction of the entire system.

After the protocol is established, the process starts with the client sending a request to the server to capture a 2D image. The server allows the Roboception camera to capture the global top-view image, and responds to the client when the task is completed. With the server's response, the client-side continues to call the deep learning network in the next step. An input image is preprocessed to remove the unnecessary regions out of the ROI by setting their pixel values to zeros. This image is fed to the deep learning network to receive a raw output. For further processing, this raw output is transferred to the Object-of-Interest Data Handling (OoIDH) module which was prepared beforehand. This module has the function to sort and group the detections which belong to the same class id wherein the recognition network provides detections in random order for each runtime even if it runs on the same input image. By using this workflow, we have overcome a limitation of PPF in detecting multiple object classes. When there are N classes of objects to be detected, instead of having to run an iterative N loops to search for the object class that has the maximum pose matching score, this pipeline, with the deep learning recognizer, could relieve this computing cost by determining upfront which class id the target belongs to and evoke PPF to estimate pose on this detection with just one run-time. In the next step, the client sends each 2D localization as bounding-box center coordinates to the server. For each center, the server performs a 2D-to-3D mapping to estimate a corresponding 3D location ($P_x, P_y, P_z$). With a given disparity image providing disparity value d, a pixel coordinates ($p_x, p_y$) i.e. the bounding-box center and a stereo baseline t, the location is readily derived by applying equation (2). Especially, the Z-depth is estimated by retrieving the min Z-depth value among a neighboring region of the input 2D center taken from the depth map. This 3D location ($P_x, P_y, P_{z_{min}}$) is the new position that the client needs to control the manipulator to move to reach the corresponding target. By then, the client has to confirm with the server whether the new position has been reached so that the server can consequently trigger the 6D pose estimator.

$$P_x = \frac{p_x \cdot t}{d} \quad P_y = \frac{p_y \cdot t}{d} \quad P_z = \frac{f \cdot t}{d} \quad (2)$$

After PPF returns the 6D pose of the target object with respect to the D435 camera, the server feedbacks this result to the client. For ground-truth calculation, we leverage the manipulator with its known pose and the fixed relationship between the camera and the end-effector which can be found by hand-eye calibration. The cycles of sending a pair of center coordinates and receiving a 6D pose repeat until the list of query bounding-box centers is empty. Next, the system can either initiate a new cycle by sending a request to capture a new image or be stopped by a request from an external device.

The general steps are summarized as followed:

| | |
|---|---|
| Step 1: | The client sends a message to the server with Msg id = 1 to request another RGB-D image capture from the server. The RGB image is used for the deep learning inference, while the corresponding depth information is later used in step 4. |
| Step 2: | By receiving a message with Msg id = 1 from client, server allows the Roboception camera to capture a new image.<br>A message with Msg id = 2 is sent to the client to confirm that the image is saved in the destination folder. |
| Step 3: | The deep learning network is executed after the client received a message with Msg id = 2, the steps are described as below:<br>✔ Preprocessing: an input image is processed to eliminate the unused regions from the ROI by setting their pixel values to zeros.<br>✔ The new image is fed to the deep learning network to get the 2D detection results including recognition and localization.<br>✔ Previous results are passed through the OoIDH to determine the number of classes, the number of instances per class, the bounding box centers, and separate them into different groups. The sorted and grouped detections are queued to a pose estimation query list.<br>The client sends a message with Msg id = 3 and the first bounding box center coordinates in the query list to the server. This message also confirms that the deep learning was successfully run on the client and provides the data for the next step. |
| Step 4: | When receiving a message with Msg id = 3 from the client, the server is activated to perform the following operations:<br>✔ Read values sent in the message.<br>✔ Using equation 2 to map the 2D pixel coordinate $(p_x, p_y)$ to the corresponding 3D location $(P_x, P_y, P_{z_{min}})$ where $P_{z_{min}}$ is found by a search for a minimizer among the Z-depth values in a neighboring region of $(p_x, p_y)$. This location guarantees the approaching step keeps a safe distance, thus enabling the D435 camera to capture |
| | the depth information of the whole target. Figure 5 depicts the calculation for the bounds of Z-depth sampling region.<br><br>A message with Msg id = 4 and the 3D location A(X, Y, $Z_{min}$) w.r.t the Roboception camera is sent to the client to confirm the task. |
| Step 5: | Following a Msg id = 4 from the server, the 3D pose estimation workflow begins. The client receives the response, calculates the new target w.r.t. the robot base's coordinates, and servo the robot to this target.<br><br>Once the manipulator has arrived, a Msg id = 4 is sent to the server from the client to confirm that the server can start the PPF. This message comes together with the target object class id. The predetermined class id helps the original PPF algorithm to avoid the N iteration loops to match each model in the pose query list with the scene point cloud for N different object classes. |
| Step 6: | PPF is activated after the server receives the message Msg id = 5. The next procedure follows:<br><br>✔ Object class ids are retrieved.<br><br>✔ Based on the object class ids, the real size of the target object is extracted to serve the point cloud removal while the target pose is estimated by referring to the corresponding 4D lookup table.<br><br>✔ Run the PPF on the processed point cloud.<br><br>✔ The output is a 4x4 matrix representing a translation and a rotation of the target object w.r.t the camera. To make the pose representation more compact, a 3x3 rotation matrix is converted to 3 Euler angles. Finally, a message Msg id = 6 along with the translation and the rotation is returned to the client. |
| Step 7: | The client receives the message Msg id = 6 and the corresponding 6D pose. The following procedure is carried out:<br><br>✔ The 6D pose is extracted. A coordinate transformation is applied to the pose to determine the final 6D pose of the target object w.r.t the robot base.<br><br>✔ Check whether this feedback is the last query in the pose query list. The subsequent step is decided upon the checking result.<br><br>❖ If NO: client sends the next bounding-box center to the server and the cycle continues to step 4.<br><br>❖ If YES: the system checks whether the user wants to stop the program. If the user allows the system to continue to run, the cycle repeats by sending a message with Msg id = 1 to the server. |

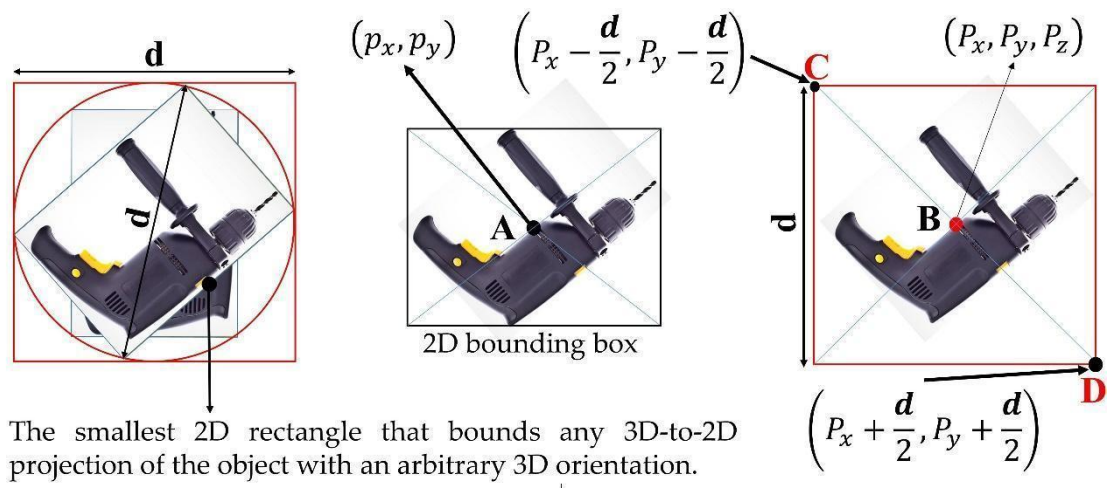

**Figure 5. The calculation of the Z-depth sampling region.** The object 2D bounding-box diameter *d* is derived as shown in the leftmost picture. This diameter *d* is used in the rightmost picture to bound the Z-depth sampling region. The image in the middle depicts the 2D bounding-box center *A* provided by the deep learning algorithm. From the 2D center *A*, a stereo 2D/3D mapping is used to compute the 3D center *B*. Finally, with *d* and *B*, a Z-depth sampling region is concretely defined by *C* and *D*.

## 4. Experiments

We divide the experiments into two stages according to each framework: instance segmentation and pose estimation and describe these stages consecutively in this section. The deep learning recognition is processed through a briefly described pipeline of data collecting, ground-truth annotation, training and testing. A similar process is used for validating the pose estimation framework. This includes an offline phase to scan and collect 3D models, preprocess point cloud data and train the model feature space. The online stage for testing follows the overall system operational pipeline as the second stage evoked by the output of instance segmentation. Our system performance is evaluated exclusively in processing time and 3D pose estimation accuracy since it can be seen that 2D recognition contributes neglectably to the pose estimation accuracy. Common pose estimation metrics require labelling of 6D pose ground-truth, we use a mechanical approach with robot calibration and a turn-table. The overall setup is shown in Figure 6:

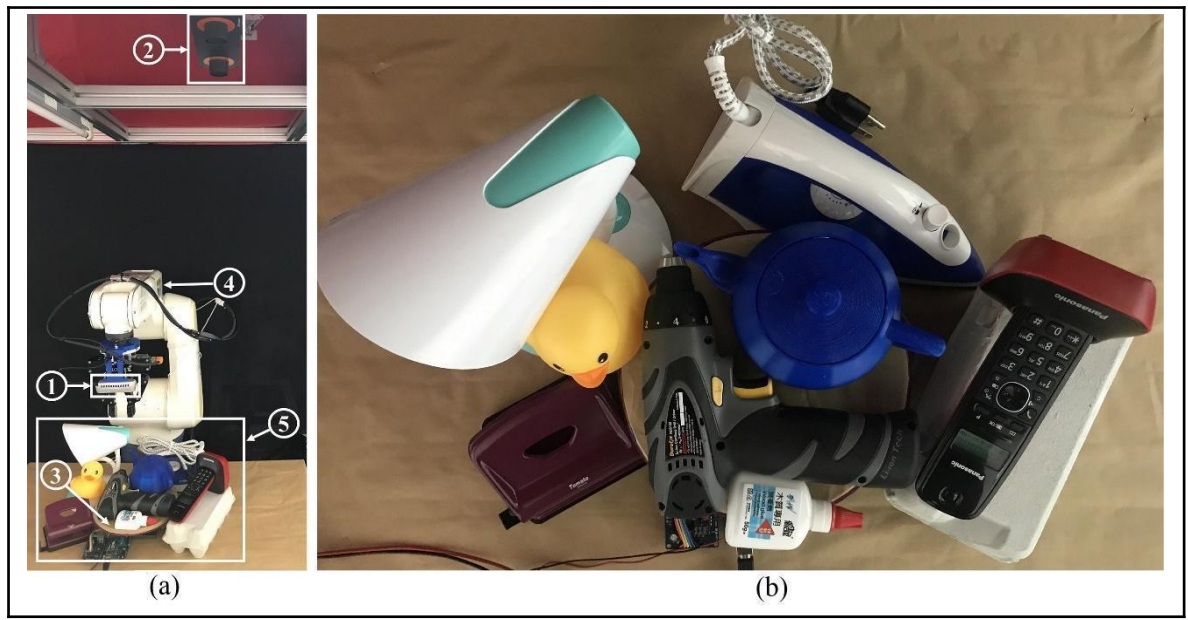

**Figure 6.** The experiment setup: (a) 1. A Realsense camera mounted in robot end-effector. 2. A Roboception camera in eye-to-hand configuration to capture top-view image in (b); 3. A turn-table to assist in object ground-truth calibration. 4. DENSO robot arm to calibrate the turn-table. 5. set of 9 objects. (b) top-view image.

Since our setup involves a second camera and a robot approach move, it is not possible to find available datasets having the same customization. This led us to develop a home-brew dataset and also collect 6D object pose labelling. This task was facilitated by a controlled turn-table shown in Figure 6a.

To quantify 3D pose estimation results, 5cm-5deg [49] and ADD [8] metrics are chosen on the basis of their popularity and straight-forward interpretation. Specifically, in the case of 5cm-5deg, when the error in translation and rotation between ground-truth and estimated pose is less than 5cm and 5deg respectively, a detection is counted as correct. In the case of the ADD metric, to compute this score, a 3D point model of the object is needed and it is computed as the average relative distance of the N model points $m_i$ when the object model is placed at ground-truth (R*, T*) and at estimated pose $(\widetilde{R}, \widetilde{T})$. A detection is considered correct if the ADD score is less than 10% of the model diameter. To calculate the ADD score, the following equation is used (3):

$$ADD = \frac{1}{N} \sum_{i=0}^{N-1} \|\left(R^* m_i + T^*\right) - \left(\widetilde{R} m_i + \widetilde{T}\right)\| \quad (3)$$

The next subsections present the experiments and results of the deep learning followed by the pose estimation frameworks.

*4.1. Deep learning*

In this section, we will explain in detail our data preparation and training of the proposed network. We also introduce our data processing technique and chosen configurations to enable the training process under the condition of limited hardware resources while preserving an optimal prediction quality. The method also reduces the processing complexity and improves the accuracy of 2D/3D mapping in the subsequent steps.

Transfer learning is used to fine-tune the neural network. A model pre-trained on the COCO dataset in a synchronized 8-GPU implementation [41] is used instead of training the network from scratch as this is a more common approach taken in recent research. In order to provide an input to 2D/3D mapping, the neural network first runs on an input image to localize the bounding-box center of image coordinates. In case the captured image does not have a similar resolution to that of the network input image, certain image processing (e.g. resizing, cropping) is needed to guarantee the network input fixed size. Again, a reverse process is applied to the network bounding-box prediction to recover their coordinates in the original image space. Such processing results in certain errors in the estimation of bounding-box locations. By conducting empirical experiments to maintain the accuracy of 2D/3D mapping, we can avoid this error by using a technique, which is explained as follows.

In an ideal scenario, the image size for training and testing is expected to be 1280x960 (see Figure 7a), which is also the full image resolution of Roboception. However, due to mediocre hardware, the system cannot train the network with input images at this resolution. We circumvent this problem by a fixed ratio cropping of the input images with a reduced size of 960x720 (see Figure 7b). In a training mode, the input image size is configured to be 960x960, while this value is changed to 1280x1280 in an inference mode.

To prepare test images for the deep learning framework, a set of 50 RGB images are collected exclusively from the Roboception camera according to our workflow, in which instance segmentation first starts to analyze the whole scene. To increase the size of the image set, an augmentation process is applied including rotation, translation and scaling, which results in a total size nearly 60 times larger than the original data.

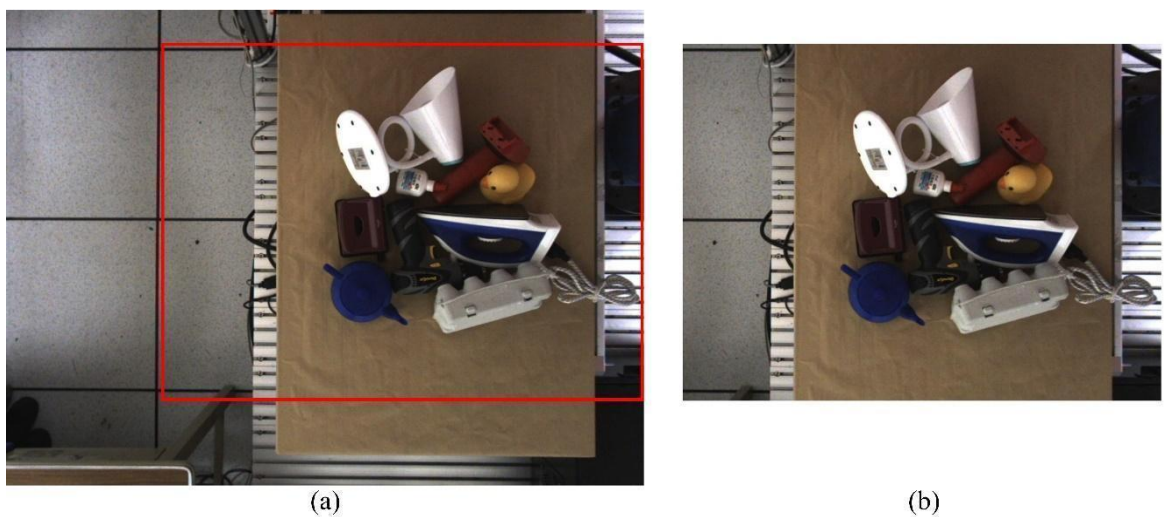

(a) (b)

**Figure 7.** A training image sample: (a) original image captured by Roboception with 1280x960 resolution; the solid-line red box designates the crop area – (b) cropped image with 960x720 resolution from (a).

In order to annotate these training images, a Python script is prepared to display an RGB input image and allow a user to define a target object boundary by mouse clicks (as shown in Figure 8). Target annotation is considered done if the start point and the end point are in a close proximity (a relative distance smaller than a predefined tolerance). The routine returns an annotation as a binary image in which the class name and the instance's IDs (within the same image) can be read by its name.

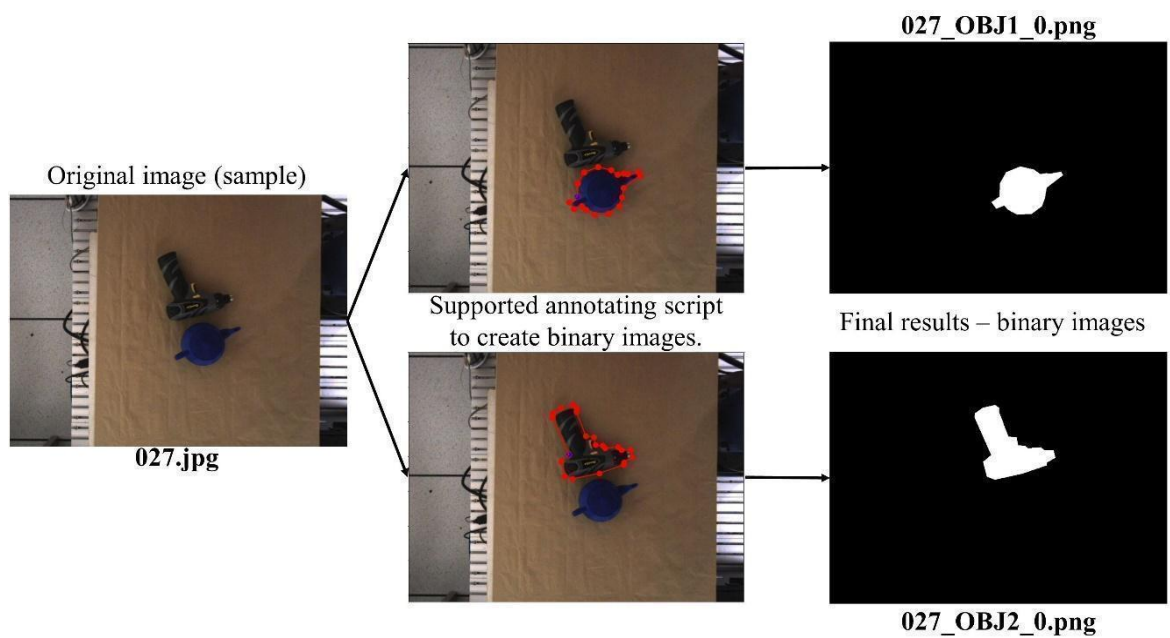

**Figure 8**. Ground-truth annotation for Mask R-CNN

Figure 9 illustrates a sample image with its augmentation. Our data augmentation scales the image at two different sizes, then rotates the image on its center. The first image from the first row is the original image, whereas its exclusive-rotation augmentation versions at a few angles are shown as the other images in the same row. On the other hand, images in the second and third rows are augmented with both scaling and rotation. The second and third row demonstrates scaling at ¾ and ½ of the original size.

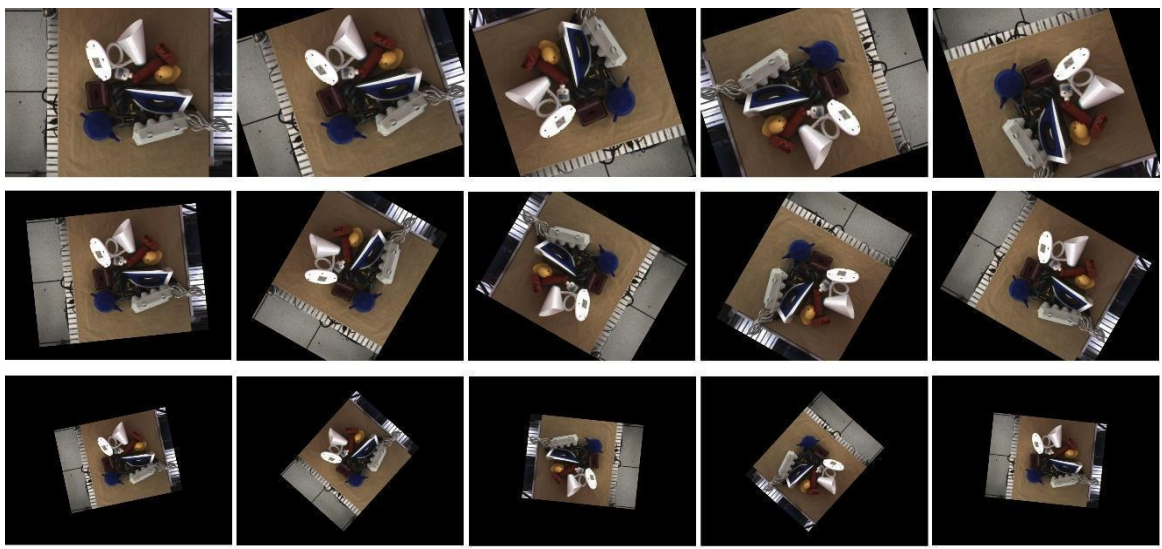

**Figure 9**. Data augmentation of the original image in the upper left with rotating and scaling.

This image set is divided to a ratio of 6:2:2 respectively for training, validation and testing. The network backbone ResNet-101 is trained a total of 20 epochs, each has 2000 iterations with an initial learning rate 0.001. There are more than 2000 images in the training set, while our model is only capable of one training image per GPU. Thus, a value of 2000 interaction per epoch can ensure that no image will be used for training more than once within an epoch. A careful consideration of this parameter can help prevent the network from overfitting. Besides, early stopping was also used during training to keep the generalization gap between training and validation loss from being widened. This can be seen in the training loss in Figure 9 where the training was stopped as soon as the losses reach a plateau point. We refer readers to [50] and [51] for more description of a similar

<span style="color:red">training method.</span>

After the initial data processing and preparation, the network is ready for training. This training involves 4 stages. In the first stage of 5 epochs, the heads and RPN are randomly initialized and trained (i.e. the pre-trained weights were not used), freezing all the backbone including the FPN layers. The next 3 stages of 12, 2, and 1 epochs are for fine-tuning where all the layers are trained with a decreasing order of magnitude of the initial learning rate by $10^0$, $10^1$, $10^2$. The hyper-parameters are set following the original paper [41], the performance of which has already been justified in previous research [39,40,52] of the object detection task.

During the training process, only positive ROIs, which have the IoU with the ground truth bounding box value larger than 0.5, are considered to contribute to the mask loss value. All the negative ones will be ignored. For each iteration, one image is used for training, and there are 256 anchors per image to use for RPN training. Each image has 200 ROIs with a ratio of 1:3 of positive to negatives. In the original paper [41], the number of sample RoIs is 512 for FPN (as in [52]). However, with a NMS threshold at 0.7 for the RPN, the RPN module cannot generate enough positive proposals to fill this and keep a positive vs negative ratio of 1:3. Therefore, we reduce this value to 200. More proposals can be achieved by adjusting the value of the RPN's NMS threshold. Following the original paper, a weight decay of 0.0001 and a momentum of 0.9 are used for our network. In our training, we set all the loss weights as equal.

In the inference mode, the proposal number is set to 1000 (as in [52]), which is used for the prediction branch to run on. These results will be checked by applying a non-maximum suppression threshold for detection at a value of 0.3. All the proposals below this threshold are skipped. Only 100 detection boxes, which obtain the highest score are used to feed into the mask graph. This setup can be seen as a way to speed up inference while improving accuracy even though it differs from the parallel computational used in the training process. Based on the predicted class k out of K classes from the classification branch, the mask branch returns the only use of the k-th mask, whereas it can provide a total of K masks per RoI.

Figure 10 presents the loss graphs of our Mask R-CNN model with data augmentation. To obtain more reliable performance, we use a stratified k-fold cross-validation with k equal 5. Figure 11 shows the results when the model runs on inference mode in four different cases.

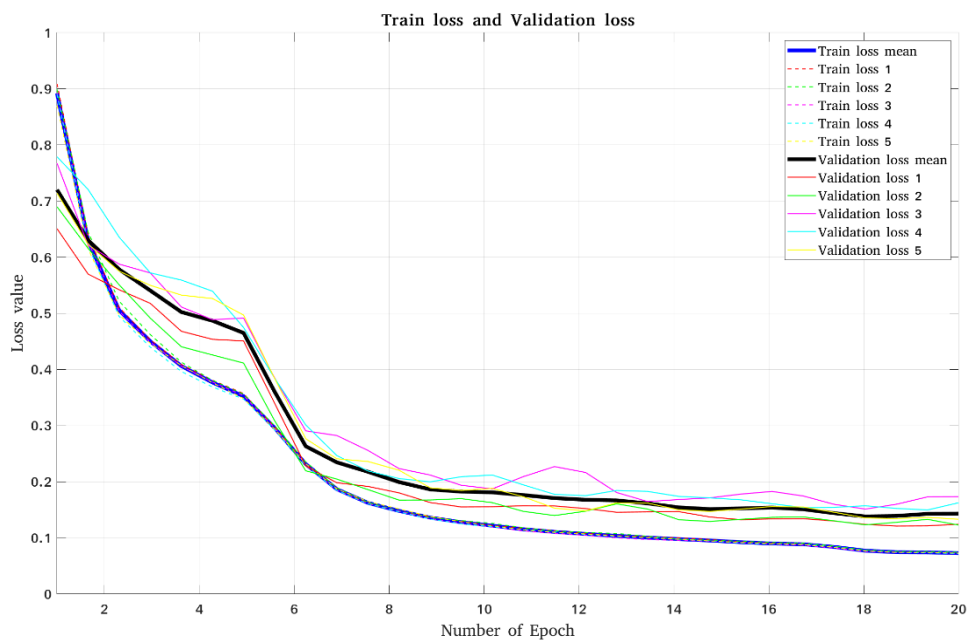

**Figure 10.** The loss graph of the deep learning model with data augmentation.

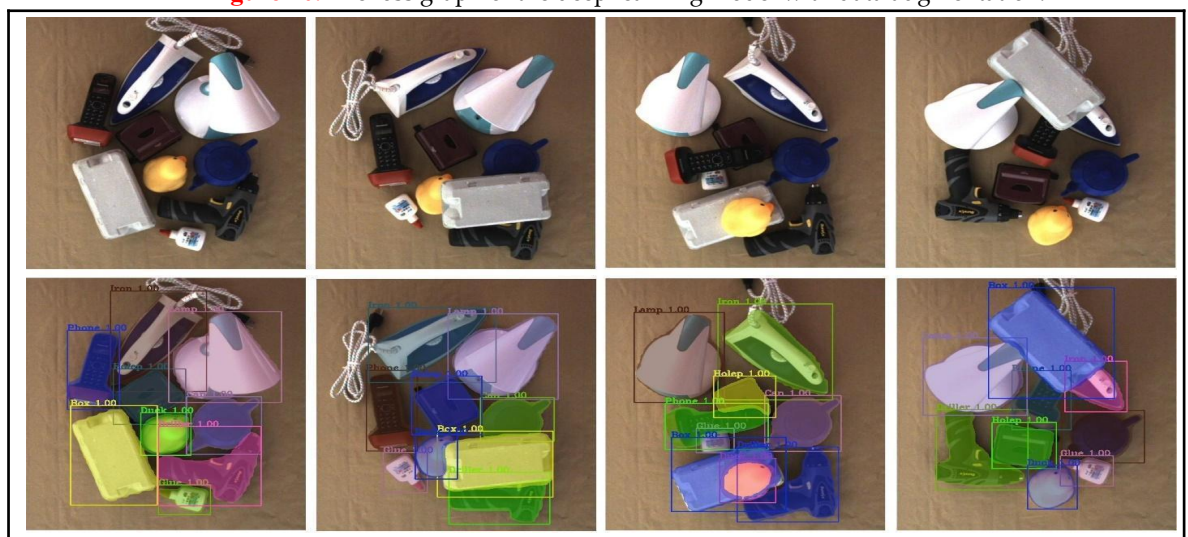

**Figure 11**. Inference running on cases: upper row shows input images, lower row shows according results.

With this setup, our tests have shown no false positive or miss detection regardless of the object class, clutter or occlusion scenarios. To coordinate with pose estimation stage, the instance segmentation detector is required to detect reliably object classes and use this result to signal the next stage. Figure 12 demonstrates this capability and our results also validated this absolute reliability of detection throughout the test set.

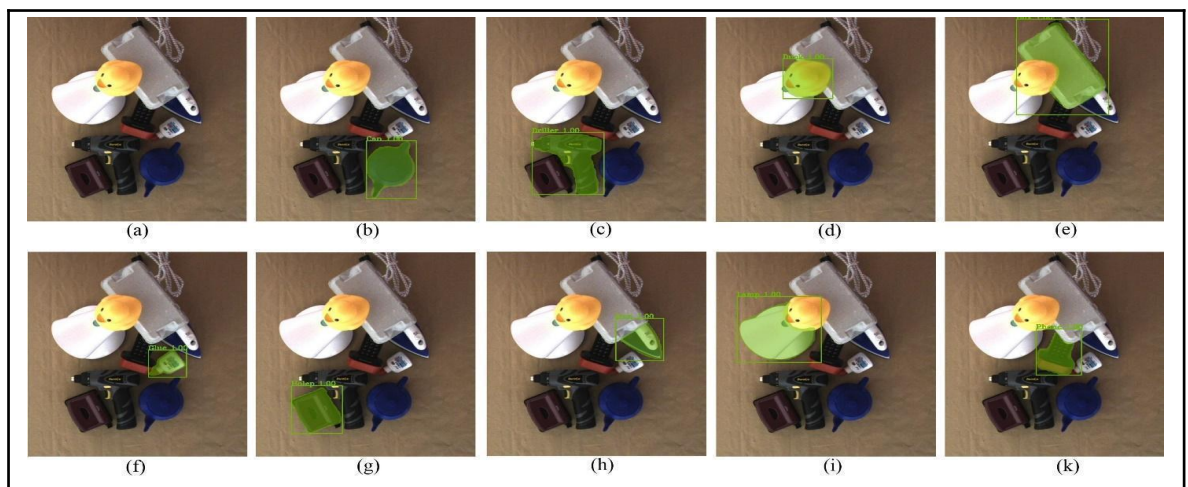

**Figure 12**. Individual object class recognition (including occluded objects): (a). input images – (b-k): detected class id = 1 to class id = 9.

*4.2. Pose estimation*

The experiment starts with the offline training by preparing the object 3D models. We use SHINING 3D EinScan-SE to construct the 3D models of 9 objects. The EinScan-SE has a single shot accuracy ≤0.1mm, a maximum scan volume up to 700×700×700mm and a scan speed < 8s. A collected object model in the form of point cloud is down-sampled to reduce the number of vertices. The processed model is used to extract features and build the hash table model space. The camera setup and result models are shown in Figure 13. The scanning of the object 3D model is one of the crucial tasks. The more similar the scanned 3D models are to the real object, the less noisy pose estimation would be during the online matching step and lead to a better performance. Therefore, each of our objects was scanned and aligned carefully to have a good 3D model. The dimensions of the object 3D bounding-box are listed in Table 1.

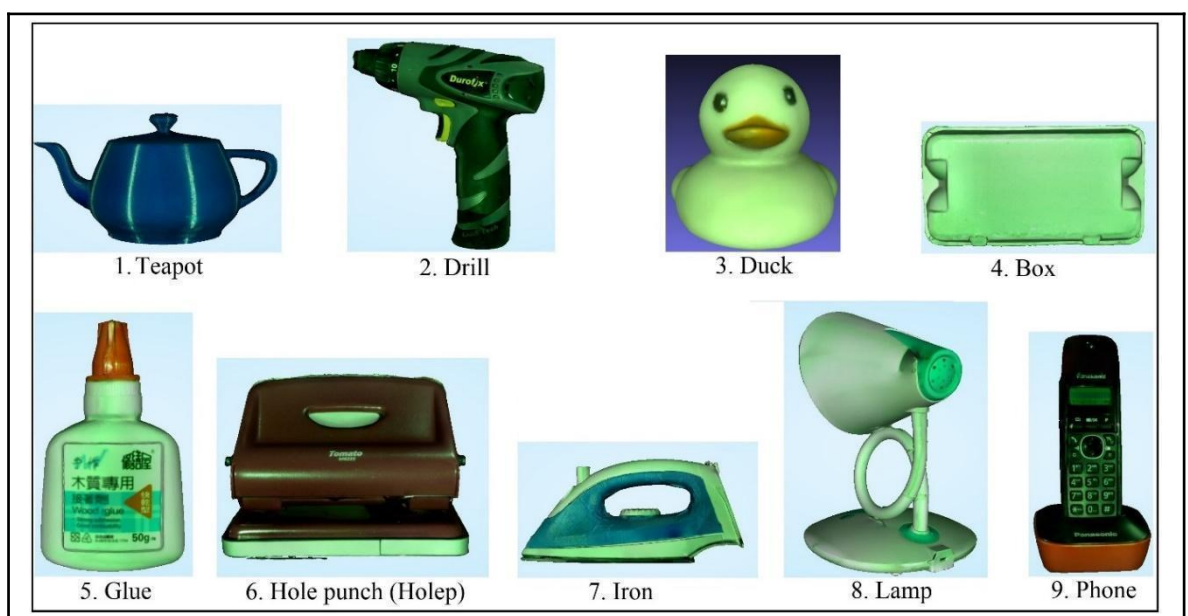

**Figure 13**. Our custom dataset of 9 objects, a comparable version of LineMod enabling comparison of our approach against other approaches benchmarked by LineMod.

**Table 1.** Dimension (mm) of the smallest object 3D bounding-box. The wood glue is the smallest sample and the lamp is the biggest sample in terms of 3D bounding-box diameter.

| Teapot | Drill | Duck | Box | Glue | Holep | Iron | Lamp | Phone |
|--------|-------|------|-----|------|-------|------|------|-------|
| 153x92x114 | 203x54x169 | 101x91x90 | 145x85x216 | 53x93x33 | 125x83x111 | 241x116x160 | 163x182x230 | 95x191x231 |

To have ground-truth for pose estimation, an additional calibration needs to be conducted. We use a mechanical approach with a robot arm and a turn-table. In brief, the robot arm with a calibrated tool-head mounted on its end-effector is moved by a teaching pendant to calibrate the table center which is used equivalently as the ground-truth location (see Figure 14). The Roboception camera is calibrated with the robot base by a utility along the software bundle. These two calibrations resolve the relative pose between the turn-table or the ground-truth and the camera is also derived.

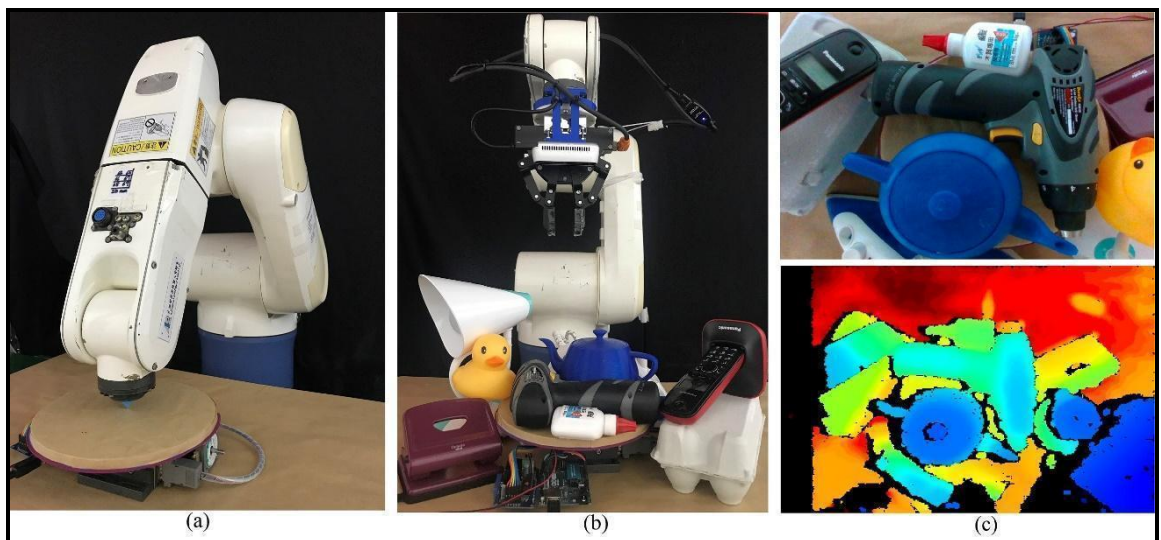

**Figure 14**. Pose estimation experiment: (a) ground-truth calibration (b) one-stage robot move before pose estimation is triggered (c) image captured from the eye-in-hand camera at the robot pose in (b).

To collect estimation results, the system online execution is used. After the object detection is provided and propagated to the second stage, the one-step approach is performed to move the robot arm to the specified target where the pose estimator is evoked. For this experiment, we created a test set of 2700 total samples, composed of the number of 30-degree rotation steps (R) of the turn-table, the number of objects (O) and the number of repetitive samples (P) at each pose: Total=R*O*P where R=12, O=9 and P=25.

To illustrate the images fed to our workflow, Figures 14 and 15 show the frames captured from the global eye-to-hand and local eye-in-hand cameras with and without occlusion, respectively. The preprocessing step removes any unnecessary regions from the ROI by setting their pixel values to zeros. Next, the deep learning network detects the target using the RGB image. Then the arm moves closer to the object and orients the D435 camera to face towards the local target. Finally, the improved PPF algorithm runs on the current point cloud and returns with the corresponding 6D estimated pose.

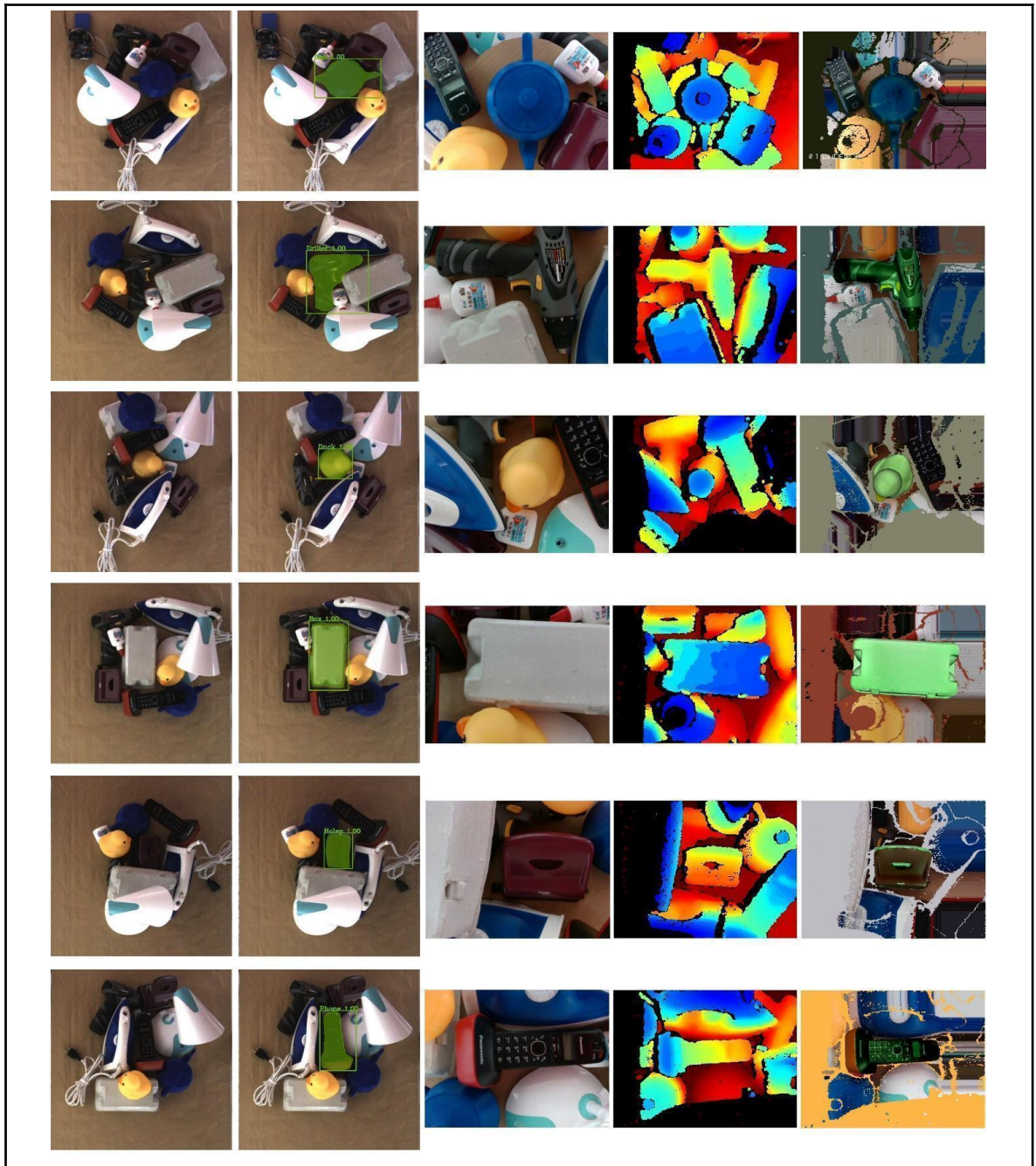

**Figure 15:** Input captured frames for the whole system in a few scenes. From left to right: 1. RGB from top-view camera – 2. Instance segmentation result – 3. RGB image (with 640x480 resolution) from the eye-in-hand D435 captured after the robot arm moved one step approach – 4. Depth image (with 320x240 resolution) – here depth image of the whole view (unapplied pointcloud removal) is used for illustration purpose – 5. Pose matching result.

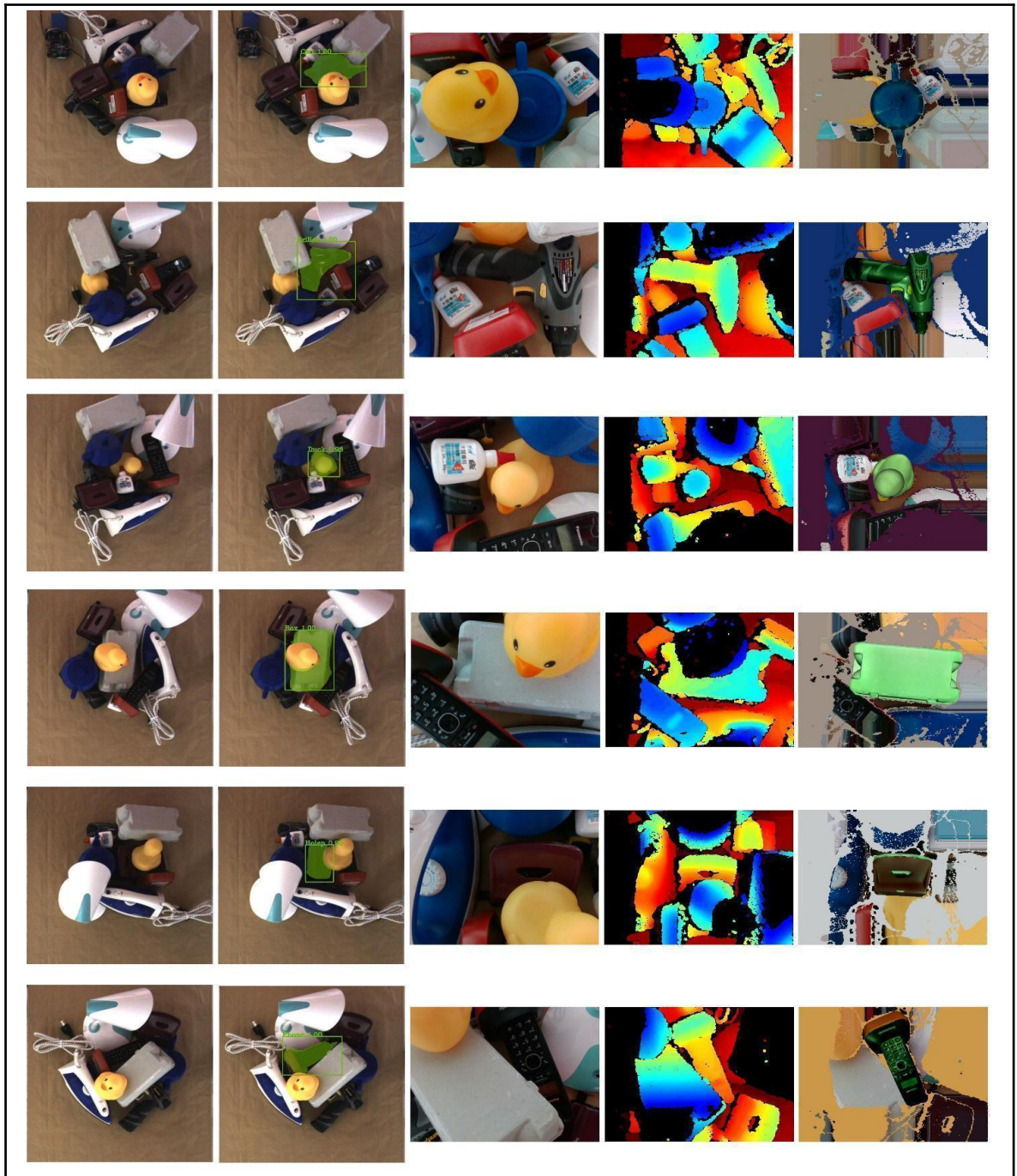

**Figure 16**. Input captured frames for the whole system in a few scenes with occlusion. From left to right: 1. RGB from top-view camera – 2. Instance segmentation result – 3. RGB image (with 640x480 resolution) from the eye-in-hand D435 captured after the robot arm moved one step approach – 4. Depth image (with 320x240 resolution) – here depth image of the whole view (unapplied pointcloud removal) is used for illustration purpose – 5. Pose matching result.

*4.3. Robotic grasping*

To demonstrate the performance in a real-world application, in this section, the proposed system is tested in a robot pick-and-place scenario. The two subjects used for this experiment are shown in Figure 17, while the real size of these two objects shows in Table 2.

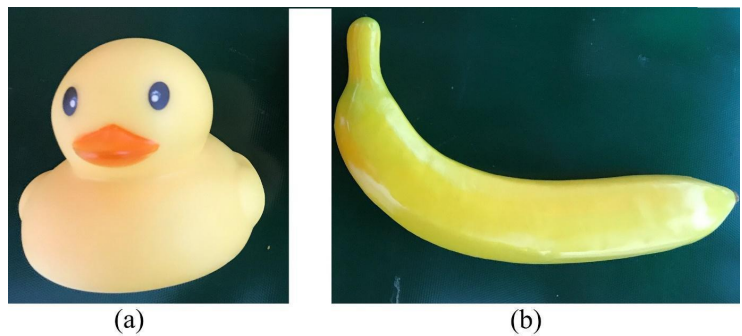

**Figure 17**. Target object for the robotics grasping experiment: a) Yellow rubber duck, b) Yellow banana.

**Table 2.** Actual dimensions of the target object.

|  | Length (mm) | Width (mm) | Depth (mm) |
|---|---|---|---|
| Yellow rubber duck | 100.61 | 91.06 | 89.58 |
| Yellow banana | 186.20 | 35.03 | 115.83 |

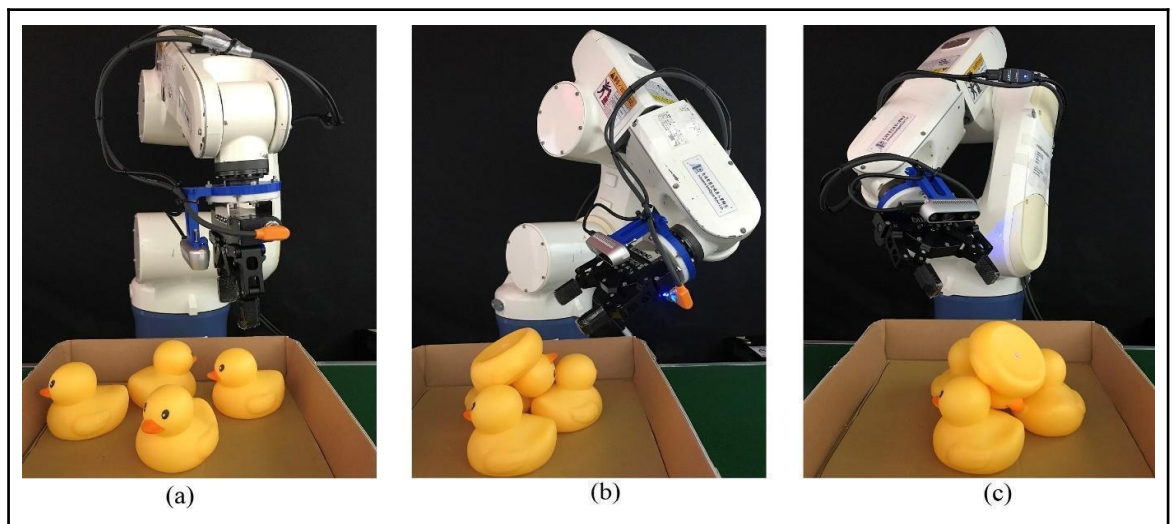

**Figure 18.** The robot's posture to grasp the objects in real-world cases: (a) camera captures the top view of the objects; (b) camera captures the front view of the objects; (c) camera captures the back view of the objects.

This test demonstrates the system's capability to handle piles of randomly arranged objects. The vision algorithms successfully estimate a target 3D pose, and the robot is controlled to grasp precisely this target despite the scene complexity. Figure 18 illustrates a few different grasping approach angles for the yellow rubber duck test.

## 5. Discussions

In this section, we present and discuss the experiment result of the items mentioned in Section 4 (4.1, 4.2, and 4.3). In previous works, the Point Pair Feature voting approach [22,53] was evaluated against the state-of-the-art solutions and outperformed these solutions on a set of extensive and publicly available datasets with a variety of challenging real-world scenarios under clutter and occlusion. Table 3. with the 2 metrics 5cm5deg and ADD have shown a clear advantage of our method compared to [22]. Especially, ADD metrics shows a high average score of 99.37%. This remarkably high score is facilitated by a closer capture distance of the point cloud resulting in less noisy signal. In addition, point cloud removal results in noise reduction in the captured point cloud. In the 5cm5deg metric, detection failure is rarely encountered i.e. 71 failures among 2700 detections. From the score of 5cm5deg, our PPF algorithm was found to have an angle estimation error closely to 5deg and this is handicapped by the metric. All these failed detections are caused by the angle estimation error. On the other hand, the ADD metric penalizes less heavily on angle error since 10% of object diameter gives a generous tolerance for errors in angle. This also shows that our pose estimator computes a rather accurate translation so that this error in combination with ~5deg angle error fit nicely under the 10% penalization of ADD metrics.

**Table 3**: Pose estimation accuracy on our dataset for single objects.

|  | Teapot | Drill | Duck | Box | Glue | Holep | Iron | Lamp | Phone | Average |
|---|---|---|---|---|---|---|---|---|---|---|
| **5cm5° metric** | | | | | | | | | | |
| Vidal et al. [22] (1.0m) | 2.67 | 36.0 | - | 39.0 | - | 0.33 | 51.67 | 27.0 | - | 17.41 |
| Vidal et al. [22] (0.85m) | 7.00 | 7.67 | 0.33 | 70.67 | - | 10.0 | **100** | 46.67 | 4.00 | 35.37 |
| Proposed method (0.2m) | **99.67** | **100** | **99.3** | **99.0** | **86.0** | **100** | **100** | **98.0** | **94.3** | **97.37** |
| **ADD metric** | | | | | | | | | | |
| Vidal et al. [22] (1.0m) | 4.67 | 93.67 | - | 56.33 | - | 2.67 | 77.67 | 55.33 | 15.0 | 33.93 |
| Vidal et al. [22] (0.85m) | 7.67 | **100** | 1.33 | 62.0 | 0.33 | 33.67 | **100** | 88.0 | 29.0 | 46.89 |
| Proposed method (0.2m) | **100** | **100** | **100** | **99.0** | **97.33** | **100** | **100** | **98.0** | **100** | **99.37** |

The report shown in Table 3 has affirmed a clear advantage when the pose estimation algorithm is applied at a closer working distance (see Figure 19). At distance 1.0 meter, the algorithm fails to estimate pose of the sample duck, glue, and phone. With just 15% closer working distance, two failed objects already can be detected. With our proposed pipeline and the one-stage move at 0.2 meter working distance, the success rate has reached convincing performance. Even the most challenging sample glue box can acquire a 86% success rate. A similar trend can be observed in the counterpart ADD metric. This one-stage move also addresses the accuracy drop for smaller objects where the algorithm can be preset its one-stage move to achieve an advantage working distance accordingly with respect to a target's size. According to the experiments, we can see a drastic change in point cloud quality when the D435 camera captures the scene at various distances and, therefore, objects with smaller size suffered severely in detection results.

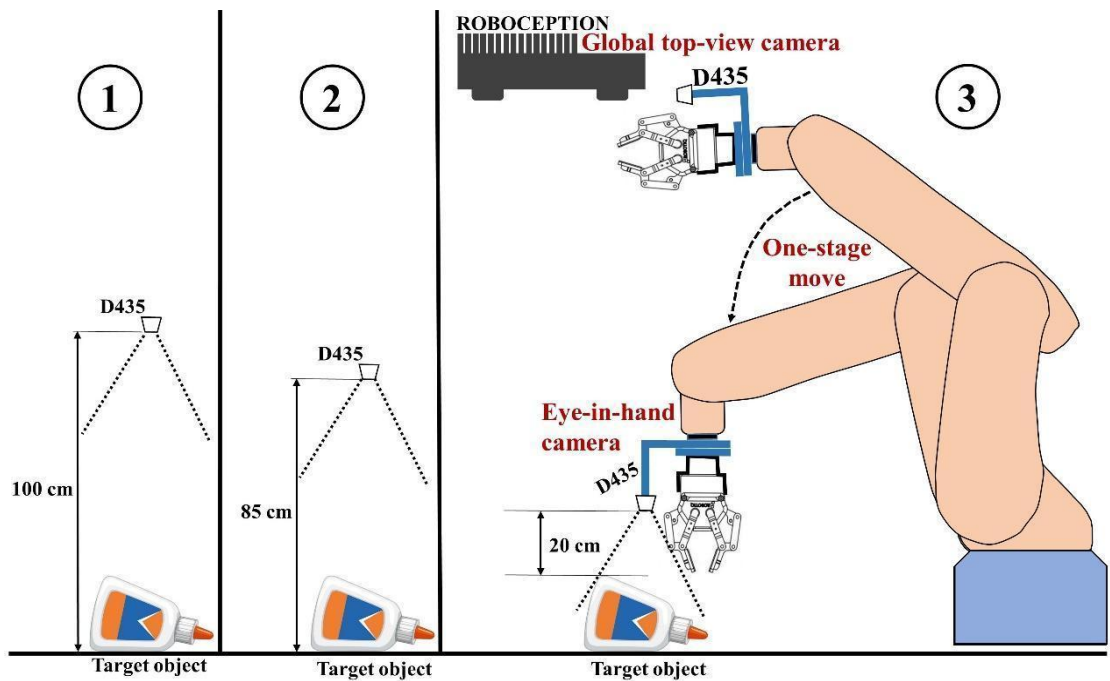

**Figure 19.** illustrates the experiment setup to reveal the advantage of the proposed workflow and system. The pose estimation algorithm was evoked at different distances: a. Vidal et al. [22] at 1.0m – b. Vidal et al. [22] at 0.85m – c. Proposed method at 0.2m (the distance is calculated relatively to the highest point among the remaining cloud points after cloud removal).

With 5cm5deg metric, little variation in metric score can be seen using the proposed method. Therefore, we re-evaluate the result with a stricter metric 3cm3deg to realize score differences among classes. Results are shown in Table 4.

**Table 4**: Pose estimation accuracy on our dataset for a single object based on 3cm3° metric.

| | Teapot | Drill | Duck | Box | Glue | Holep | Iron | Lamp | Phone | Average |
|---|---|---|---|---|---|---|---|---|---|---|
| Proposed method (0.2m) | 85.0 | 67.0 | 91.3 | 99.0 | 50.3 | 99.6 | 99.3 | 98.0 | 70.7 | 84.47 |

It can be seen that the score differences between classes are significant, especially plummet for the cylindrical or elliptical cylindrical shape objects like glue, driller, phone. As mentioned earlier, 5cm5deg metric penalizes the cases where the angle error is greater than 5deg. By carefully inspecting detection results of these objects. we have found that the angle estimation error varies in a range 2 ~ 3.5 degree. Therefore, with this stricter metric, many previously passed cases become handicapped where angle estimation error is greater than 3 degree.

In addition to the metrics, we also collect distribution graphs to demonstrate the system accuracy. The samples in the graph are drawn from the success detection with respect to 3cm3deg metrics (see Figure 20). The distribution covers 3 translations, 3 rotations for all the object samples. In both rotation and translation, estimation along the z-axis is the most accurate. Among the samples, *teapot* exhibits the smallest translation error whereas the biggest one can be seen in *lamp*. Translation error has a wider value range from approximately 1mm (*teapot*) to ~8mm (*lamp*). In contrast, rotation error is more stably contained in the 3deg range.

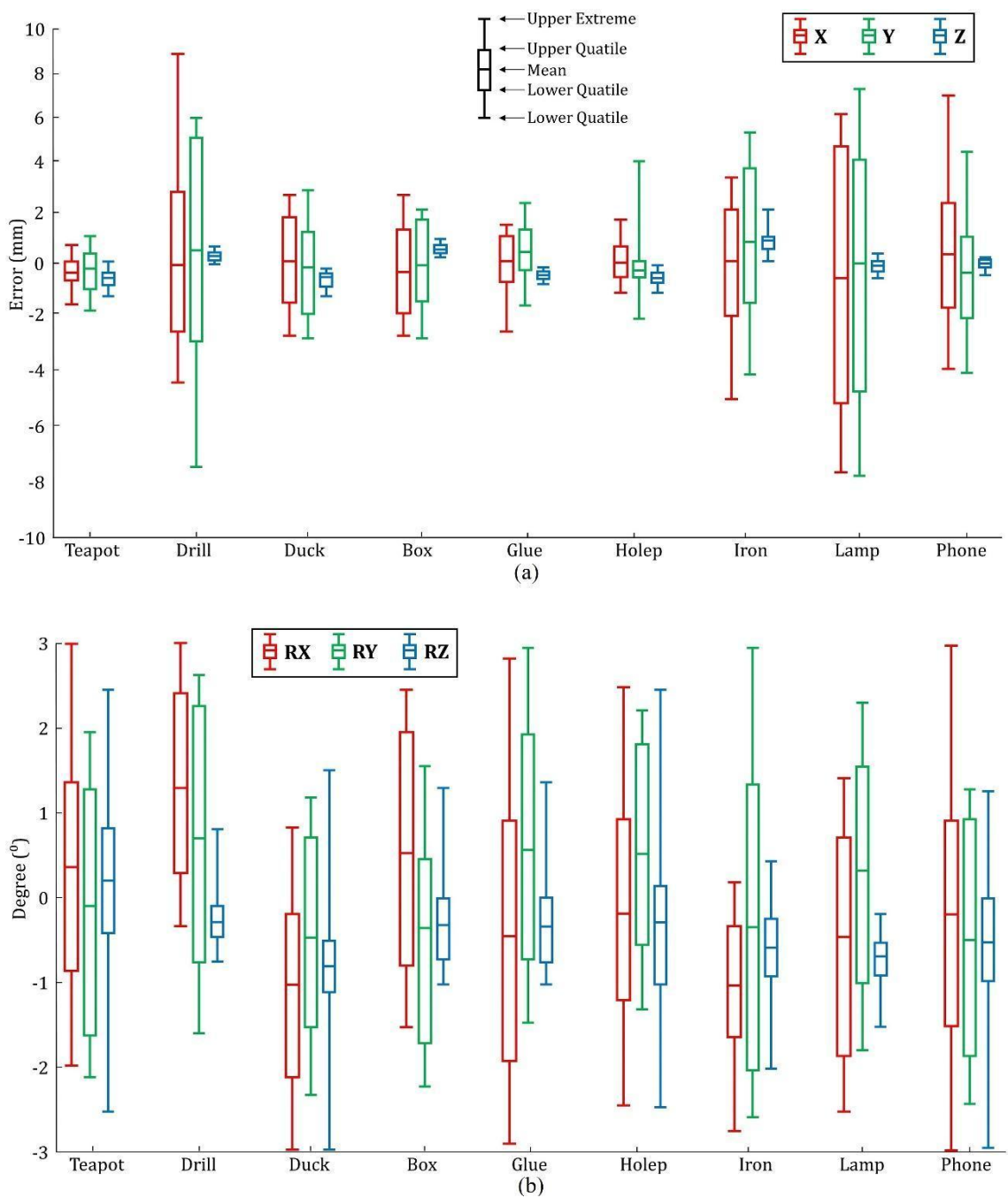

**Figure 20.** Error distribution for all samples: a) translation X-Y-Z; b) rotation RX-RY-RZ.

Besides the accuracy limits as shown with the two metrics, some other detection failures are about to be discussed. Figure 21 showed a test case with the glue object. Particularly, this object is nearly symmetric (except the embossed characters KS on one side Figure 21e). For this case, in addition to the small size, the object is also occluded partially. This results in an inconsistent pose detection in Figure 21d (correct detection) or Figure 21e (wrong detection to a flipped of the object).

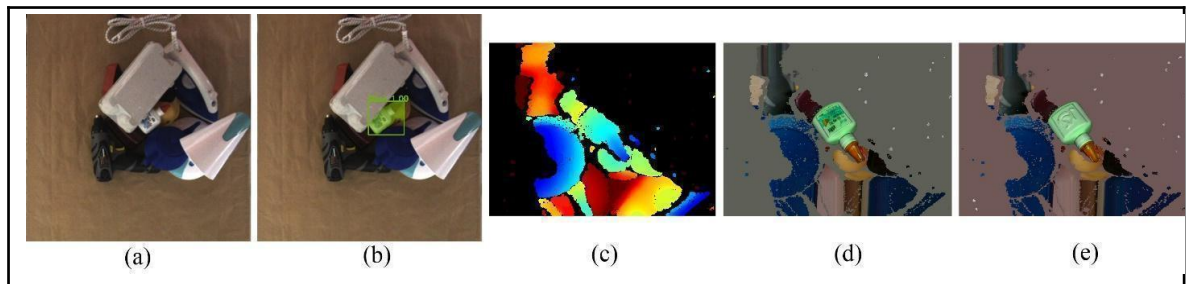

**Figure 21**. Failure cases with symmetric objects: detection result is not consistent. However, depending on applications, all symmetrical sides can be considered as a correct detection.

Another common failure is the insufficiently discriminative information in the depth map. In the following test, the setup is intentionally prepared to investigate this failure. Obviously, there are cases where the occlusion removes or replaces a high portion of the 3D model. The rest point cloud may be too few to estimate an accurate pose. Figure 22's first row depicts such scenarios. In this particular case, the iron is mostly occluded and remains a few visible depth information (Figure 22c). This results in an inconsistent prediction including correct (Figure 22d) and wrong one (Figure 22e)

The last failure scenario is self-occlusion. The second row in Figure 22 shows the lamp object having this scenario. The depth image can capture only the lamp head which already occludes its body and base. Moreover, this head has a symmetric shape and results in a wrong prediction (Figure 22k). In such a challenging case, there are still correct detections Figure 22i. However, this result is not stable.

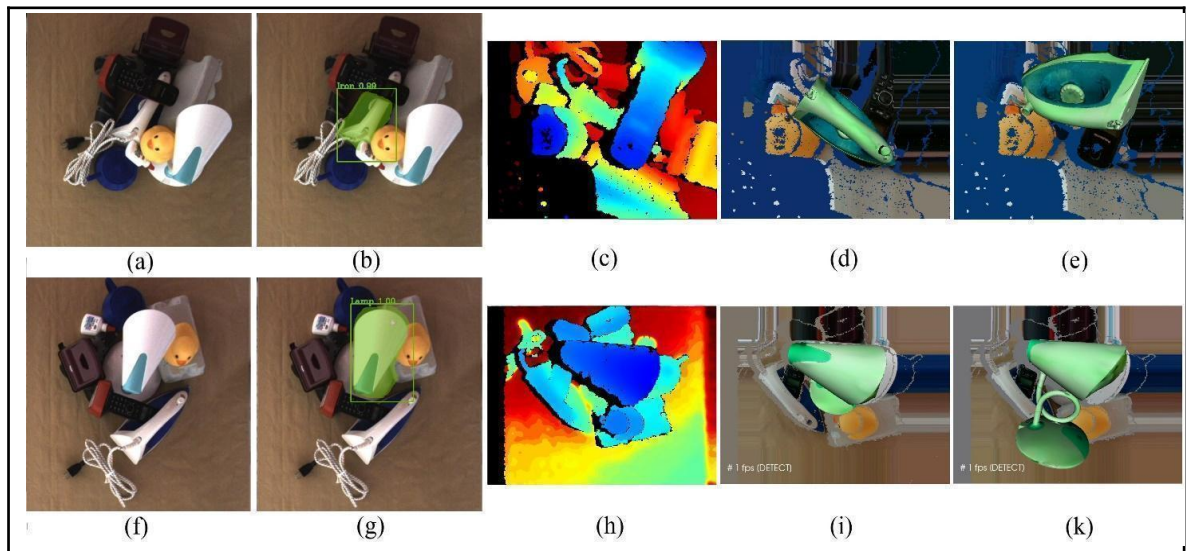

**Figure 22**. depth information is not sufficient for a precise pose estimation (two scenarios) - Scenario 1: the iron occlusion - the sample is occluded by other objects. Scenario 2: the lamp self-occlusion - the occlusion results in a partially visible object (i.e. only the lamp head is visible and it occludes the lamp body and base). Such scenarios may lead to unstable pose estimation which may rapidly alternate between correct estimation (21d,i) and erroneous ones (21e,k).

Table 5 has shown the run-time comparison between the Vidal et al. [22] pose estimation and our proposed method. The run-time at each row is collected by 50 measurements for each object whereas the first and second row indicate the information for the PPF algorithm without our proposed one-stage move and with the move respectively. It can be noted that PPF run-time on an object varies depending on object dimension, target, and scene surface features. In particular, run-time of the proposed method takes, in average, 52.4% of the original method. This is explained by the point cloud removal facilitated by the deep learning prediction. However, since this one-stage move is preset at a fixed distance to different target objects regardless of their size, this does not fully take advantage of the point cloud removal. For example, the improvement of sample *lamp* can be seen as little as 8.57%.

Table 5. Run-time comparison between Vidal et al. [22] and our proposal method. The average, biggest, and smallest run-time improvements are 47,6%, 73.88% (iron), and 8.51% (lamp), respectively.

|  | Teapot | Drill | Duck | Box | Glue | Holep | Iron | Lamp | Phone | Average |
|---|---|---|---|---|---|---|---|---|---|---|
| Vidal et al. [22] (ms) | 1044 | 1315 | 933 | 2394 | 831 | 1220 | 3018 | 1257 | 1322 | 1482 |
| Proposed method (ms) | 757 | 848 | 657 | 731 | 614 | 682 | 788 | 1150 | 760 | 776 |

The robot performed 60 attempts in total and the final report is shown in Table 6. Among all the tests in Section 4.3, the failures were mostly caused by the gripper. The 2-Finger Gripper was too big for close contact with the target in cluttered scenes, as the fingers collided with the surrounded objects during close approach to the target. This altered the target's position before the gripper successfully closed its fingers while grasping.

Table 6. Pickup success rate using 2-Finger Adaptive Robot Gripper (Robotiq)

|  | Total trials | Success | Fail | Success rate (%) |
|---|---|---|---|---|
| Yellow rubber duck | 30 | 26 | 4 | 86.67 |
| Yellow banana | 30 | 28 | 2 | 93.33 |
| Average | 60 | 54 | 6 | 90 |

With the motivation to bridge the gap to practical applications, we plan to continue testing our proposed system with a bigger variety of real scenarios. This involves variation in object appearance, materials, and arrangement. This expects to reveal the vulnerability of the system for future development direction. To effectively improve the grasping performance, a soft robotics gripper or a vacuum suction pad end-effector could replace the 2-Finger Gripper in future work. Lastly, additional work can be done to achieve an active vision capability that can automatically adjust the camera-object working distance accordingly to object sizes to take the most advantage out of the cloud removal stage.

6. Conclusions

In this research, we have proposed a 3D object recognition solution that inherits two contemporary state-of-the-art frameworks including a neural network based 2D recognizer and a feature-based 3D pose estimator. Our solution is efficiently capable of 3D object recognition in multi object class scenarios and can achieve more precise pose estimation by leveraging the one-stage move workflow. In order to meet the robust performance requirement by practical applications, we additionally introduce a two-step workflow that utilizes a global top-view camera for initial deep learning detection. This first detection supervises the one-stage robot move bringing the eye-in-hand camera closer to the target ready for the next capture and processing before evoking pose estimation. To coordinate these two frameworks efficiently, we design and implement a custom client-server interface between them. Our intensive tests were conducted for both constituent frameworks. Experiment results have shown a consistent reliability of our system on a variety of test scenarios including occlusion cases. Specifically, our solution has achieved a higher score in 5cm-5deg metric and ADD metric among several other competitive methods. Finally, a pick-and-place robotics experiment is conducted to illustrate the overall efficiency of the proposed system in a real-world application, and has shown a convincing success rate.

**Author Contributions:** Conceptualization, Chyi-Yeu Lin; Data curation, Tuan-Tang Le and Trung-Son Le; Formal analysis, Tuan-Tang Le; Funding acquisition, Chyi-Yeu Lin; Investigation, Tuan-Tang Le and Trung-Son Le; Methodology, Tuan-Tang Le, Trung-Son Le, Yu-Ru Chen and Joel Vidal ; Project administration, Chyi-Yeu Lin; Software, Tuan-Tang Le and Joel Vidal ; Supervision, Chyi-Yeu Lin; Validation, Tuan-Tang Le; Visualization, Tuan-Tang Le; Writing – original draft, Tuan-Tang Le and Trung-Son Le; Writing – review & editing, Joel Vidal and Chyi-Yeu Lin.

**Funding:** This work is financially supported by both Taiwan Building Technology Center and Center for Cyber-physical System Innovation from the Featured Areas Research Center Program within the framework of the Higher Education Sprout Project by the Ministry of Education (MOE) in Taiwan.

**Acknowledgments:** This research was financially supported by the Ministry of Science and Technology, Republic of China, under the grant 105-2221-E-011 -088 -MY2.

**Conflicts of Interest:** The authors declare no conflict of interest.

32 of 33object pose estimation in cluttered scenes, ArXiv Prepr. ArXiv1711.00199. (2017).

[44]  S. Peng, Y. Liu, Q. Huang, X. Zhou, H. Bao, Pvnet: Pixel-wise voting network for 6dof pose estimation, in: Proc. IEEE Conf. Comput. Vis. Pattern Recognit., 2019: pp. 4561–4570.

[45]  W. Kehl, F. Manhardt, F. Tombari, S. Ilic, N. Navab, Ssd-6d: Making rgb-based 3d detection and 6d pose estimation great again, in: Proc. IEEE Int. Conf. Comput. Vis., 2017: pp. 1521–1529.

[46]  W. Abdulla, Mask R-CNN for object detection and instance segmentation on Keras and TensorFlow, GitHub Repos. (2017).

[47]  W. Liu, D. Anguelov, D. Erhan, C. Szegedy, S. Reed, C.-Y. Fu, A.C. Berg, Ssd: Single shot multibox detector, in: Eur. Conf. Comput. Vis., 2016: pp. 21–37.

[48]  J. Redmon, S. Divvala, R. Girshick, A. Farhadi, You only look once: Unified, real-time object detection, in: Proc. IEEE Conf. Comput. Vis. Pattern Recognit., 2016: pp. 779–788.

[49]  J. Shotton, B. Glocker, C. Zach, S. Izadi, A. Criminisi, A. Fitzgibbon, Scene coordinate regression forests for camera relocalization in RGB-D images, in: Proc. IEEE Conf. Comput. Vis. Pattern Recognit., 2013: pp. 2930–2937.

[50]  T.-T. Le, C.-Y. Lin, E.J. Piedad, Deep learning for noninvasive classification of clustered horticultural crops – A case for banana fruit tiers, Postharvest Biol. Technol. 156 (2019). https://doi.org/10.1016/j.postharvbio.2019.05.023.

[51]  T.-T. Le, C.-Y. Lin, Bin-Picking for Planar Objects Based on a Deep Learning Network: A Case Study of USB Packs, Sensors. 19 (2019) 3602.

[52]  T.-Y. Lin, P. Dollár, R. Girshick, K. He, B. Hariharan, S. Belongie, Feature pyramid networks for object detection, in: Proc. IEEE Conf. Comput. Vis. Pattern Recognit., 2017: pp. 2117–2125.

[53]  J. Vidal, C.-Y. Lin, R. Mart\'\i, 6D pose estimation using an improved method based on point pair features, in: 2018 4th Int. Conf. Control. Autom. Robot., 2018: pp. 405–409.

[55]  Yu-Ru Chen, Ping-Hsuan Wu, Chyi-Yeu, Lin and Yuan-Chiu. "Efficient pose estimation strategy for household object grasping with 2D and 3D vision." 16th Int. Conf. on Automation Technology, 2019.

[56]  Yu-Ru Chen, . "Robot Arm Autonomous Object Grasping System Based on 2D and 3D Vision Techniques." Master's thesis, National Taiwan University of Science and Technology, 2018.